\documentclass[pmlr]{jmlr}

\usepackage{longtable}

\usepackage{booktabs}
\usepackage{adjustbox}
\usepackage{multirow}
\usepackage[load-configurations=version-1]{siunitx} 

\makeatletter
\def\set@curr@file#1{\def\@curr@file{#1}} 
\makeatother

\theorembodyfont{\upshape}
\theoremheaderfont{\scshape}
\theorempostheader{:}
\theoremsep{\newline}

\title[Multimodal Representation Learning of CMR]{Multimodal Representation Learning of Cardiovascular Magnetic Resonance Imaging}

\author{\Name{Jielin Qiu$^{*1}$} \Email{jielinq@andrew.cmu.edu}  \\
\Name{Peide Huang$^{*1}$} \Email{peideh@andrew.cmu.edu} \\
\Name{Makiya Nakashima$^2$} \Email{nakashm2@ccf.org} \\
\Name{Jaehyun Lee$^2$} \Email{leej62@ccf.org} \\
\Name{Jiacheng Zhu$^1$} \Email{jzhu4@andrew.cmu.edu} \\
\Name{Wilson Tang$^2$} \Email{tangw@ccf.org} \\
\Name{Pohao Chen$^2$} \Email{chenp2@ccf.org} \\
\Name{Christopher Nguyen$^2$} \Email{nguyenc6@ccf.org} \\
\Name{Byung-Hak Kim$^3$} \Email{byunghakk@gmail.com} \\
\Name{Debbie Kwon$^2$} \Email{kwond@ccf.org} \\
\Name{Douglas Weber$^1$} \Email{dougweber@cmu.edu} \\
\Name{Ding Zhao$^1$} \Email{dingzhao@cmu.edu} \\
\Name{David Chen$^2$} \Email{chend3@ccf.org} \\
\addr $^1$ Carnegie Mellon University\\
$^2$ Heart Vascular and Thoracic Institute, Cleveland Clinic \\
$^3$ CJ AI Center
\AND
\footnotemark[1] \addr {\normalfont \footnotesize marked as equal contribution}
}

\begin{document}

\maketitle

\begin{abstract}
Self-supervised learning is crucial for clinical imaging applications, given the lack of explicit labels in healthcare. However, conventional approaches that rely on precise vision-language alignment are not always feasible in complex clinical imaging modalities, such as cardiac magnetic resonance (CMR). CMR provides a comprehensive visualization of cardiac anatomy, physiology, and microstructure, making it challenging to interpret. Additionally, CMR reports require synthesizing information from sequences of images and different views, resulting in potentially weak alignment between the study and diagnosis report pair.
To overcome these challenges, we propose \textbf{CMRformer}, a multimodal learning framework to jointly learn sequences of CMR images and associated cardiologist's reports. 
Moreover, one of the major obstacles to improving CMR study is the lack of large, publicly available datasets. To bridge this gap, we collected a large \textbf{CMR dataset}, which consists of 13,787 studies from clinical cases. 
By utilizing our proposed CMRformer and our collected dataset, we achieved remarkable performance in real-world clinical tasks, such as CMR image retrieval and diagnosis report retrieval.
Furthermore, the learned representations are evaluated to be practically helpful for downstream applications, such as disease classification.
Our work could potentially expedite progress in the CMR study and lead to more accurate and effective diagnosis and treatment.
\end{abstract}

\section{Introduction}

The application of deep learning to clinical imaging is a highly researched field given the extensive potential to alleviate overburdened providers through automation \citep{hosny_artificial_2018}, improve medical care through standardization \citep{nagendran_artificial_2020}, and accelerate research through high knowledge discovery \citep{dash_big_2019}. However, many current applications are hindered by the lack of annotated data imposed by high costs \citep{lee_big_2017} and privacy regulations \citep{good_crowdsourcing_2013}. Recent innovations in self-supervised learning where using pretext tasks or implied knowledge have provided a method for which to reduce the dependence on large annotated datasets. Many proposed frameworks are constrained to a single domain, such as image or text \citep{azizi_big_2021, bai_self-supervised_2019, chen_self-supervised_2019, he_masked_2022}. Yet, this would ignore the natural association between clinical images and written radiology reports containing expert interpretations. 

Recent developments in contrastive image-text pretraining \citep{radford_learning_2021} leverage the natural alignment between image and text pairs to provide co-supervision for each domain and achieved good performance on the natural image-text pairs collected from Internet \citep{Chen2020UNITERUI,Gan2020LargeScaleAT,li2022unimo,Li2020OscarOA,Kim2021ViLTVT,Li2021AlignBF,Li2022BLIPBL,Dou2021AnES}. However, clinical imaging has unique properties not often seen in natural images. Cardiac magnetic resonance imaging (CMR) is one such example. CMR allows users to visualize the 3D cardiac anatomy and function in an unlimited number of views (although standardized to a few American Heart Association-specified views \citep{schulz-menger_standardized_2020}). CMR studies are able to visualize the morphology, motion, tissue characteristics, and even tissue perfusion within a single study. Each type of image has different characteristics, which make them sensitive to different pathophysiologies. As such, the associated radiology report incorporates findings that describe both individual images and findings that synthesize from multiple image types and views. 

\begin{figure}[t]
\centering
\includegraphics[width=0.99\textwidth]{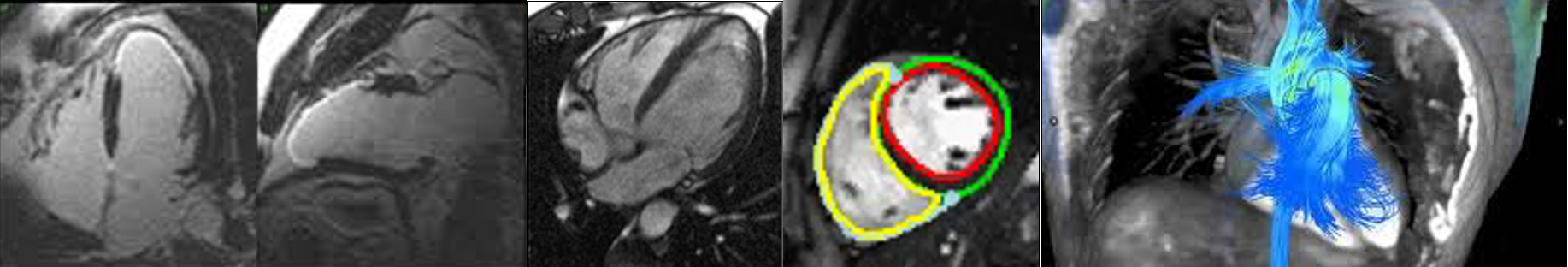}
\caption{Examples of Cardiac magnetic resonance (CMR) images. CMR comprises hundreds of images of differing views and image types, the large majority of which do not contain an indication of pathology. The corresponding report is an analysis of all these images, resulting in difficulty in aligning images with text reports.}
\vspace{-20pt}
\label{fig:cmr}
\end{figure}

Existing image-text pretraining in the natural image domain assume significant alignment between a single image and a single piece of text. For instance, the natural image MS-COCO  dataset \citep{lin_microsoft_2014} uses crowd-sourced short blurbs to describe each image. This simplistic formatting is reflected in the current application of image-text pretraining in clinical imaging domains. Previous works within the clinical imaging field have mostly focused on chest X-ray \citep{irvin_chexpert_2019, johnson_mimic-cxr_2019, wang_chestx-ray8_2017}, where the radiology reports are often of limited length, and the image domain is limited to only one or two views of the subject. Furthermore, despite the relatively high resolution of chest X-ray images, the saliency area is often fairly prominent in each image.
In contrast, it is not straightforward to incorporate CMR images into current vision-language frameworks, given that only a few images within possibly thousands contained in a CMR study have visible pathology. For example, a ventricular septal defect is a serious morphological abnormality that often requires invasive surgery. However, the defect may be visible in only one or two slices among dozens acquired. Explicitly aligning the written interpretation with individual images is difficult, given that interpretation involves synthesizing images from the whole study, leading to poor alignment between individual images and text.

To address the issue encountered in CMR studies, we propose \textbf{CMRformer}, a multimodal learning framework, to jointly learn visual features from CMR images and textual features from radiology reports. Specifically, the CMR data are structured as a sequence of images in video format. Our contribution can be summarized as follows:
\vspace{-5pt}
\begin{itemize}
    \item We propose \textbf{CMRformer}, a multimodal learning framework that learns visual features from CMR images and textual features from text reports in a joint manner to address the issue of weak alignment between CMR image sequences and their corresponding diagnosis textual reports. The framework is designed to learn from the entire CMR study without the need for manual identification of specific images relevant to particular diseases.
    \vspace{-5pt}
    \item The learned embeddings hold immense potential for practical clinical applications, such as the retrieval of CMR studies or radiology reports, searching and retrieving specific information from vast amounts of data in a more efficient manner.
    \vspace{-5pt}
    \item As the existing literature falls short in providing a large public dataset for CMR, we took the initiative to gather a comprehensive dataset consisting of 13,787 studies derived from actual clinical cases. We also collected and labeled a Cardiomyopathies dataset for the downstream classification task, which included 1,939 Cardiomyopathies studies and was labeled by clinical experts.
\end{itemize}

\subsection*{Generalizable Insights about Machine Learning in the Context of Healthcare}

In this work, we address three fundamental problems existing for clinical CMR imaging learning: (1) the lack of adequate volume of data for individual clinical tasks as well as access to expert labels; (2) the weak alignment between CMR images and the associated reports, which is the bottleneck for multimodal learning; and (3) the lack of functional model which can account for the weak alignment of CMR data to learn useful embeddings for downstream clinical applications. 

\vspace{5pt}
Our work addresses the problems above by:
\begin{itemize}
    \item We collected a large, single-site CMR dataset consisting of 13,787 studies derived from actual clinical cases. We also collected and labeled a Cardiomyopathies dataset, with 1,939 studies for the downstream disease classification task.
    \vspace{-5pt}
    \item We proposed CMRformer, a multimodal learning framework that enables the learning of weak alignments between CMR images and corresponding doctor's reports. The potential applications of this framework are many-fold and far-reaching, including applications in content-based information retrieval, clinical decision support, and healthcare operations. We believe that it represents a significant step forward in the field of medical image analysis and clinical decision-making for CMR study.
    \vspace{-5pt}
    \item Although this framework was used on CMR data in our study, we expect it can be generalized across several clinical imaging modalities and other healthcare data, including echocardiography, computed tomography, and multi-omic data for the more robust and generalizable application of deep learning to the healthcare domain. Our framework facilitates the extraction of valuable insights from disparate sources of medical data, empowering clinicians with the ability to make more informed and accurate diagnoses and treatment decisions. Ultimately, the widespread adoption of this framework has the potential to significantly improve patient treatment.
\end{itemize}

\section{Related Work}


\paragraph{Medical Multimodal Learning in Image-Text Setting}
Medical multimodal learning in the image-text setting focuses on learning the alignment between medical images and accompanying text using a contrastive image-text learning framework. \citet{zhang_contrastive_2022} proposed ConVIRT, a contrastive image and text self-supervised learning framework similar to simCLR for chest X-rays, which exceeded the supervised end-to-end method with only $1 \%$ of the training data. GLoRIA \citep{huang_gloria_2021} introduced a cross-attention layer in order to learn localized similarities in both image and word subdomains. Similarly,  LoVT \citep{muller2022radiological}  leveraged a projector for localized representations.
Recently, \citet{wang_medclip_2022} proposed to leverage prior knowledge (Unified Medical Language System \citet{bodenreider_unified_2004}) as distant supervision to the contrastive learning process. \citet{yan2022clinical} proposed combining contrastive learning with mask language modeling to train ClinicalBERT. \citet{windsor2023vision} explored various data augmentation strategies to improve data efficiency in the clinical realm. However, these methods highly depend on strong alignment between image and text pairs.

\paragraph{Multimodal Learning in Video-Text Setting}
In video-language pretraining, most clips are not semantically well aligned with their corresponding text \citep{miech_howto100m_2019, miech20endtoend}. For example, “the basketball player makes a game-winning shot” may have several seconds of additional gameplay and celebration for context. \citet{xu_videoclip_2021} varied the lengths of the video clips and enforced overlapping clips to drive increased similarity between closely related clips, which work for instructional videos or video captioning where there is still a strong assumption of alignment between the action and the text. 
Alternatively, \citet{buch_revisiting_2022,bain_frozen_2021} identified that single frames within videos contribute vast amounts of information. Specifically, \citet{buch_revisiting_2022} leveraged an image-based embedding with a self-attention mechanism to identify the most informative frame for any specific piece of text. \citet{bain_frozen_2021} proposed a space-time transformer-based encoder that can take both video and image jointly to learn the importance of spatial and temporal features, minimizing the need for well-aligned video-text pairs.

\section{Methods}
\begin{figure}[htp]
\centering
\includegraphics[width=0.9\textwidth]{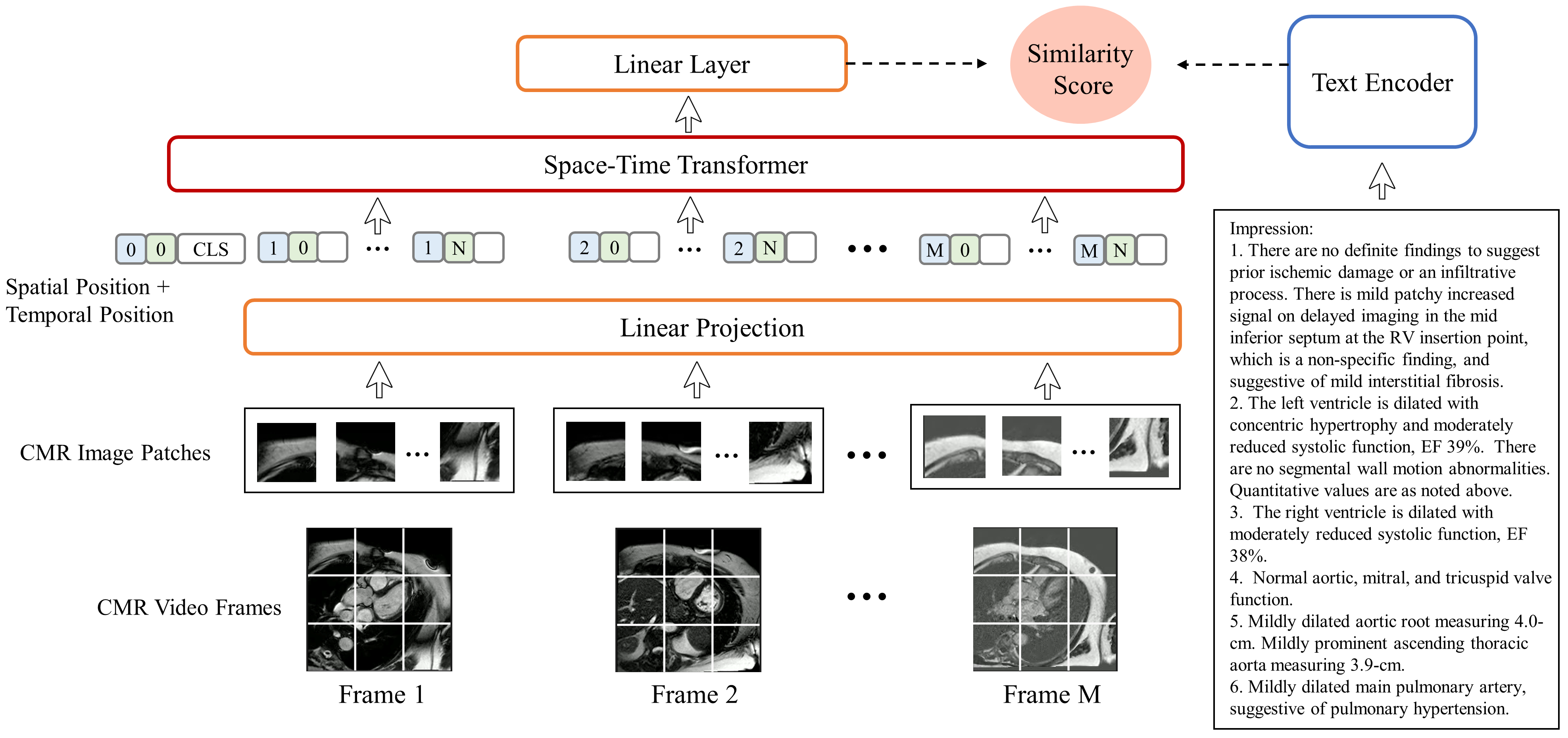}
\caption{The overall architecture of our model, where the visual encoder processes sequences of CMR images and the text encoder processes the text from the ``impression" section of the corresponding reports.}
\label{fig:model}
\vspace{-10pt}
\end{figure}

To learn better CMR-report multimodal representations, we proposed \textbf{CMRformer}, a multimodal learning framework, to jointly learn visual features from CMR studies and text embeddings from the associated radiologists' reports. The model architecture is shown in Figure~\ref{fig:model}, which contains a visual encoder to process CMR images, and a text encoder to process text reports. More details are introduced in the following sections.

\subsection{Model Architecture}

\paragraph{Visual encoder} 
The visual encoder processes an image or video clip $X \in \mathrm{R}^{M \times 3 \times H \times W}$, where $M$ is the number of frames (1 for images) with a resolution of $H \times W$. It comprises three main components: (i) the patch embedding layer, (ii) learnable embeddings for positional space, time, and [CLS], and (iii) a stack of 12 space-time attention blocks.

To generate patch embeddings, the patch embedding layer uses a 2D convolutional layer with a kernel and stride size equal to the target patch size $P=16$, with $d=768$ output channels (the chosen embedding dimension for the video encoder). The positional space and time embeddings have shapes $M \times d$ and $N \times d$, respectively, where $M$ is the maximum number of input video frames, and $N$ is the maximum number of non-overlapping patches of size $P$ within a frame (196 for a video resolution of $224\times224$). The [CLS] embedding has shape $1 \times d$.
Each space-time attention block includes norm layers, temporal and spatial self-attention layers, and a MLP layer, following the approach described in \citet{Bain2021FrozenIT}.

To process spatio-temporal patches, the video clip input is divided into non-overlapping patches of size $P \times P$, following the protocol in ViT and Timesformer~\citep{bertasius2021spacetime}. The resulting patches $\boldsymbol x \in \mathrm{R}^{M \times N \times 3 \times P \times P}$, where $N=HW/P^2$, are passed through a 2D convolutional layer, and the output is flattened, generating a sequence of embeddings $\boldsymbol z \in \mathrm{R}^{MN \times D}$ that is fed into the transformer. The size of the embeddings, $D$, depends on the number of kernels in the convolutional layer.

To account for the temporal and spatial position of the patches, learned temporal and spatial positional embeddings, $\boldsymbol E^{s} \in \mathrm{R}^{N \times D}$ and $\boldsymbol E^{t} \in \mathrm{R}^{M \times D}$, are added to each input token as follows:
\begin{equation}
{\boldsymbol z^{(0)}{p,m}} = \boldsymbol z{p,m} + \boldsymbol E^{s}{p} + \boldsymbol E^{t}{m},
\end{equation}
In this equation, all patches within the same frame $m$ (but different spatial locations) receive the same temporal positional embedding $E^{t}{m}$, and all patches in the same spatial location (but different frames) receive the same spatial positional embedding $E^{s}{p}$. This approach enables the model to identify the temporal and spatial position of each patch. Additionally, a learned [CLS] token~\citep{devlin2019bert} is concatenated at the beginning of the sequence to produce the final visual embedding output of the transformer.

The video sequence is processed by a stack of space-time transformer blocks. We introduce a slight modification to the Divided Space-Time attention method proposed by~\citep{bertasius2021spacetime}, replacing the residual connection between the block input and the temporal attention output with a residual connection between the block input and the spatial attention output. Each block applies temporal self-attention and then spatial self-attention sequentially to the output of the previous block. Finally, the video clip embedding is derived from the [CLS] token of the final block.

\paragraph{Text encoder} 

The text encoder component of our architecture is responsible for processing a sequence of tokenized words and producing a meaningful encoding that captures the semantic content of the input text. To achieve this, we employ a multi-layer bidirectional transformer encoder, which has demonstrated remarkable performance in a wide range of natural language processing tasks~\citep{devlin2019bert,Radford2018ImprovingLU}. Specifically, we instantiate the text encoder using the \texttt{distilbert-base-uncased} model~\citep{distilbert}, which is a variant of BERT~\citep{devlin2019bert} that has been optimized for efficiency by reducing the number of layers by a factor of 2 and removing the token-type embeddings and pooler components.

At a high level, the text encoder processes the tokenized input sequence by iteratively transforming the embeddings of each token based on its context within the sentence, using a series of self-attention and feed-forward layers. The self-attention mechanism enables the model to attend to different parts of the input sequence when processing each token, while the feed-forward layers allow for nonlinear transformations of the learned representations. Moreover, the bidirectional nature of the encoder allows the model to incorporate information from both past and future tokens in the sequence, resulting in a more comprehensive encoding of the input text.

To obtain the final text encoding, we extract the output of the special [CLS] token in the final layer of the text encoder. This token is specifically designed to provide a summary representation of the input sequence that can be used for downstream tasks such as text classification or information retrieval. The text encoder component plays a crucial role in our architecture by enabling the model to incorporate textual information that complements the visual information encoded by the video encoder.

\paragraph{Projection} In order to establish a meaningful association between textual and visual information, it is imperative to first ensure that they are represented in a common feature space. To achieve this, both the textual and visual encodings are projected onto a shared dimension using separate linear layers. Subsequently, the similarity between these projected embeddings is computed by taking their dot product. This approach effectively enables the alignment of heterogeneous modalities, namely textual and visual, and facilitates their comparison in a manner that can be meaningfully interpreted by downstream tasks.

\paragraph{Efficiency} The employed model incorporates independent dual encoder pathways, as seen in the MIL-NCE~\citep{miech20endtoend} and MMV networks~\citep{alayrac2020self}, which necessitate a mere dot product computation between the video and text embeddings for establishing meaningful associations between the two modalities. The aforementioned design choice confers upon the model the advantage of a retrieval inference of trivial computational cost, as it can be indexed and efficiently queried using fast approximate nearest neighbor search methods, making it amenable to scaling for very large-scale retrieval tasks at inference time. Specifically, for a given target gallery consisting of $v$ videos and $t$ text queries, the retrieval complexity of our model is $O(t+v)$. By contrast, the ClipBERT~\citep{lei2021less} model, which adopts a single encoder for both text and video inputs, exhibits a significantly higher retrieval complexity of $O(tv)$, as every possible text-video combination needs to be inputted into the model. Other retrieval methods, such as MoEE~\citep{miech18learning} and MMT~\citep{gabeur2020multi}, which are based on expert models, also incorporate dual encoder pathways. However, they require query-conditioned weights to calculate similarity scores for each expert, which results in higher computational complexity.

\subsection{Learning Objectives}

\paragraph{Training loss}We adopt the approach introduced in~\citet{Zhai2019ClassificationIA} for a retrieval-based setting, where pairs of text and video data points in a batch are considered as positive matches, while all other pairwise combinations in the batch are considered as negative samples. To facilitate this, we minimize the sum of two losses, namely video-to-text and text-to-video, given by:
\begin{equation}
L_{v2t} = -\frac{1}{B}\sum_i^B\log{\frac{\exp(x_i^\top y_i / \sigma)}{\sum_{j=1}^{B} \exp(x_i^\top y_j / \sigma)}},~~
L_{t2v} = -\frac{1}{B}\sum_i^B\log{\frac{\exp(y_i^\top x_i / \sigma)}{\sum_{j=1}^{B} \exp(y_i^\top x_j / \sigma)}}
\end{equation}
Here, $x_i$ and $y_j$ correspond to the normalized embeddings of the $i$-th video and $j$-th text, respectively, in a batch of size $B$, while $\sigma$ denotes the temperature parameter. The video-to-text loss, $L_{v2t}$, computes the negative log probability of each video embedding matching its corresponding text embedding, relative to all other text embeddings in the batch, whereas the text-to-video loss, $L_{t2v}$, computes the negative log probability of each text embedding matching its corresponding video embedding, relative to all other video embeddings in the batch. By minimizing these losses, the model learns to generate embeddings that maximize the similarity between the corresponding video-text pairs and minimize the similarity between non-corresponding pairs.

\paragraph{Frame sampling} To capture the temporal information in a video, we divide it into $M$ segments with equal duration, where each segment contains $L/M$ frames. During training, we apply a uniform frame sampling strategy to obtain one frame from each segment. This approach, which is similar to TSN~\citep{wang_tsn} and GST~\citep{gst}, ensures that the model can learn to recognize and capture the salient features across different time segments of the video.
At inference time, we adopt a more fine-grained sampling approach by extracting the $i$-th frame from each segment, where $i$ is determined by a predefined stride $S$. This results in an array of video embeddings $\boldsymbol v = [v_0, v_{S}, v_{2S}, ..., v_{(M-1)S}]$, each of which captures a distinct temporal segment of the video. We then compute the mean of these embeddings, which provides a compact representation of the entire video while preserving the temporal information.
This approach allows the model to capture both the short-term and long-term temporal dynamics of the video and can effectively encode the information needed for downstream tasks such as video classification and retrieval. By sampling frames at different intervals during training and testing, our model can learn to recognize temporal patterns at different scales, resulting in robust and accurate representations of the video content.

\begin{figure}[t]
\centering
\includegraphics[width=0.9\textwidth]{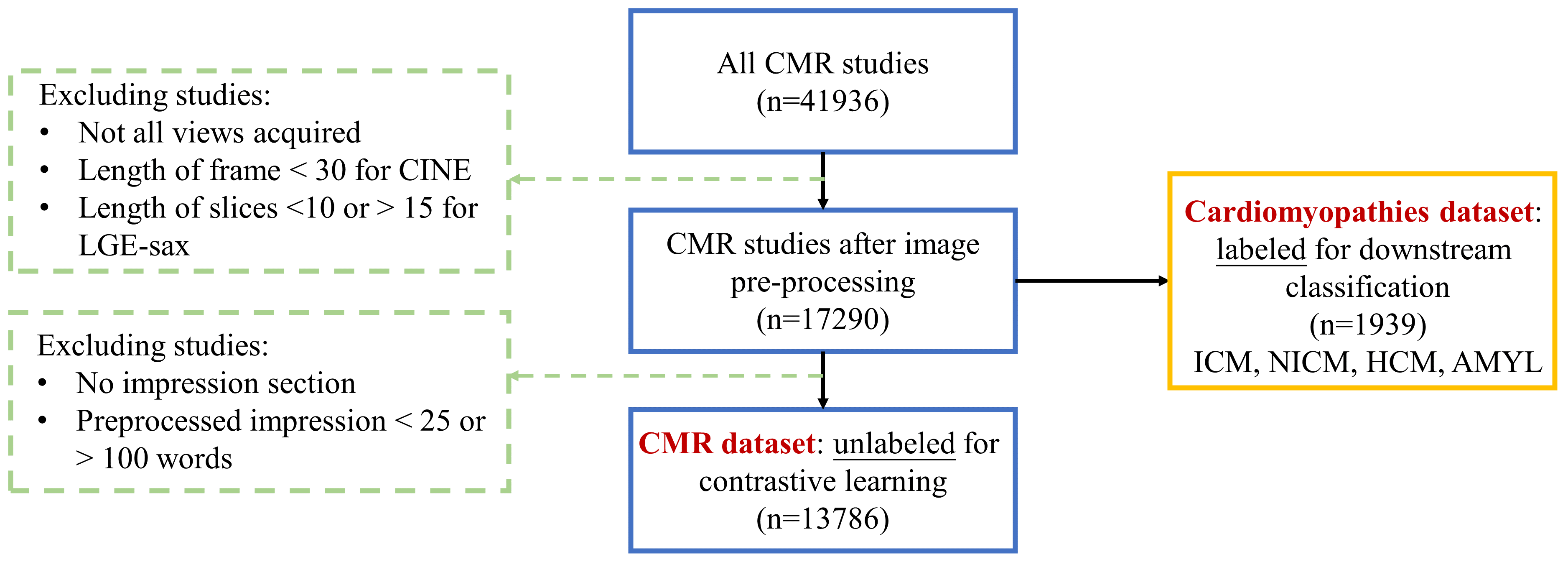}
\caption{Data prepossessing pipeline of our CMR dataset.}
\label{fig:preprocessing}
\end{figure}

\section{Our CMR Dataset}

Due to no large publicly available CMR dataset suitable for our study, we collected a \textbf{CMR dataset} by ourselves. 
Our dataset contains CMR images and cardiologists' text reports of patients who underwent a CMR exam at \textbf{Anonymous institution} in both inpatient and outpatient settings between 2008 and 2022. The total size of the initial cohort is 41,936 CMR studies, including a variety of indications according to standard clinical practices. The dataset is collected under a wide range of protocols and machines that evolved with changing clinical standards. We introduced two datasets in our study, one is the \textbf{CMR dataset}, which will be used for training our CMRformer model and the retrieval task, and the other one is the \textbf{Cardiomyopathies dataset}, which will be used in the downstream classification task. The \textbf{CMR dataset} will be introduced in this section, and the \textbf{Cardiomyopathies dataset} will be introduced in Section~\ref{sec:classification-Cardiomyopathies} in the experiment section.

\subsection{Specific Characteristics of the CMR Data}
CMR studies differ from chest X-rays and cardiac computed tomography in terms of data type. While the latter two provide single-plane or 3D views of the heart, CMR studies comprise hundreds of 2D images from various angles and image types, with some images being static while others need to be combined to form short video clips. Each image type and view convey unique information. To address this complexity, we use image metadata and established CMR interpretation standards to construct a dictionary that utilizes the ``series" DICOM header field, enabling us to specify the cardiac view and image type accurately. 

\subsection{Data Preprocessing and Filtering}
The CMR dataset preprocessing pipeline is shown in Table \ref{fig:preprocessing}, which includes the inclusion/exclusion criteria. The vast majority of cases were collected by Phillips 1.5T Achieva scanner or 3.0T Ingenia scanner. We identified studies that comprised a standard CMR exam targeting the ventricular disease. 

\paragraph{CMR image types} The two image types included in this study are: (1) \textbf{CINE}, which are high quality, 2D image clips to capture motion in a single slice of the heart; and (2) \textbf{LGE}, which standards for late gadolinium enhancement, a static 2D image which captures tissue viability. Multiple images of a single type are acquired at different slice locations to capture the whole volume of the heart. 

\paragraph{CMR image views} We target four standard CMR views, including (1) 4 chamber long axis (\textbf{lax}), which aims at looking at all 4 chambers of the heart in one view; (2) short axis (\textbf{sax}), which aims at visualizing the ventricles; (3) 2 chamber long axis (\textbf{2ch}), which aims at left heart visualization; and (4) 3 chamber long axis (\textbf{3ch}), which aims at aortic outflow track visualization. 

In this work, we define the different settings as IMAGE-TYPE$_{view}$, where IMAGE-TYPE is the type of the image, and ${view}$ is the view of the image. The studies without these scans were typically abbreviated study protocols targeting the either aortic or valvular disease.

\subsection{CMR Text Reports}
The radiology report is an important medico-legal document that contains multiple information of an imaging study to the referring clinician \citep{kahn2009toward}. As such, the report often contains technical information about the exam, the clinical history of the patient, a general list of imaging biomarkers and other findings, and summary findings split into the \textit{technique}, \textit{indications}, \textit{findings}, and \textit{impressions} section respectively. Much of this information is extraneous to the specific study, making it difficult to learn from, i.e., tabular imaging measurements in the findings section. Therefore, we target the \textbf{``impression''} section, which contains a summary of the key findings of each CMR study.

\subsection{Statistics of the Data} 
We provide a quantitative analysis of the distribution of the CMR dataset, including the length of the preprocessed texts from the ``impression" section, and the number of images from each study. The results are shown in Table~\ref{table:text_length_stats} and Table~\ref{table:image_num_stats}, respectively. Based on the tables, we could find that most CMR study contains 400-500 images, and the length of words within the ``impression" section in the corresponding reports is mostly around 30-80. For more demographics of the CMR dataset, the age range is $54.87\pm15.91$, and within the 13,786 patients, 6,133 are female and 7,653 are male.

\begin{table*}[t]
\centering
\parbox{.44\linewidth}{
\caption{Statistics of length of impression sections from text reports.}
\vspace{-15pt}
\begin{center}
\begin{adjustbox}{width=0.86\linewidth}
\begin{tabular}{c|c|c}
\toprule 
Text Length & Count & Percentage\\
\midrule
20-30 & 728 & 5.3\% \\
30-40 & 2445 & 17.7\% \\
40-50 & 2926 & 21.2\% \\
50-60 & 2666 & 19.3\% \\
60-70 & 2078 & 15.1\% \\
70-80 & 1425 & 10.3\% \\
80-90 & 929 & 6.7\% \\
90-100 & 589 & 4.3\% \\
\bottomrule
\end{tabular}
\end{adjustbox}
\end{center}
\label{table:text_length_stats}
}
\hfill
\parbox{.49\linewidth}{
\caption{Statistics of the number of images of each study.}
\vspace{-15pt}
\begin{center}
\begin{adjustbox}{width=0.9\linewidth}
\begin{tabular}{c|c|c}
\toprule 
Number of Images & Count & Percentage\\
\midrule
0-200 & 61 & 0.4\% \\
200-300 & 24 & 0.2\% \\
300-400 & 1166 & 8.5\% \\
400-500 & 12054 & 87.4\% \\
500-600 & 126 & 0.9\% \\
600-700 & 79 & 0.6\% \\
700-800 & 162 & 1.2\% \\
$>800$ & 114 & 0.8\% \\
\bottomrule
\end{tabular}
\end{adjustbox}
\end{center}
\label{table:image_num_stats}
}
\vspace{-10pt}
\end{table*}

\label{table:demographics}

\subsection{Comparison with Existing CMR Datasets} 

We compare our dataset with several existing public CMR datasets with greater than 100 included studies, including Automated Cardiac Diagnosis Challenge (ACDC) \citep{bernard2018deep}, Kaggle 2nd Annual Data Science Bowl Cardiac Challenge (DSB-CC) \citep{kaggle_dsbcc}, and Statistical Atlases and Computational Modeling of the Heart (STACOM) \citep{fonseca2011cardiac}. 
In Table~\ref{table:data_comp}, it can be observed that our dataset consists of a notably greater number of studies as compared to the datasets currently available. Furthermore, our dataset is unique in that it is directly paired with radiologist-interpreted reports.
More information about existing datasets can be found in Appendix~\ref{sec:Appendix-datasets}.

\begin{table}[t]
\centering
\caption{Comparison with existing CMR datasets.}
\begin{center}
\begin{adjustbox}{width=0.99\linewidth}
\begin{tabular}{l|c|c|c}
\toprule 
Source & Studies & Image Types & Labels\\
\midrule
ACDC  & 150 & Cine & segmentation \\
DSB-CC  & 1,140 & Cine & end-systolic and end-diastolic volumes\\
STACOM  & $<$200 & varies (mostly Cine) & varies (mostly segmentation)\\
Ours & 13,786/1,939 & Cine, LGE & radiology reports/cardiomyopathy diagnosis\\
\bottomrule
\end{tabular}
\end{adjustbox}
\end{center}
\label{table:data_comp}
\vspace{-10pt}
\end{table}

\section{Experiments}

\subsection{Experimental Setting}

\begin{figure}[t]
\centering
\includegraphics[width=0.99\textwidth]{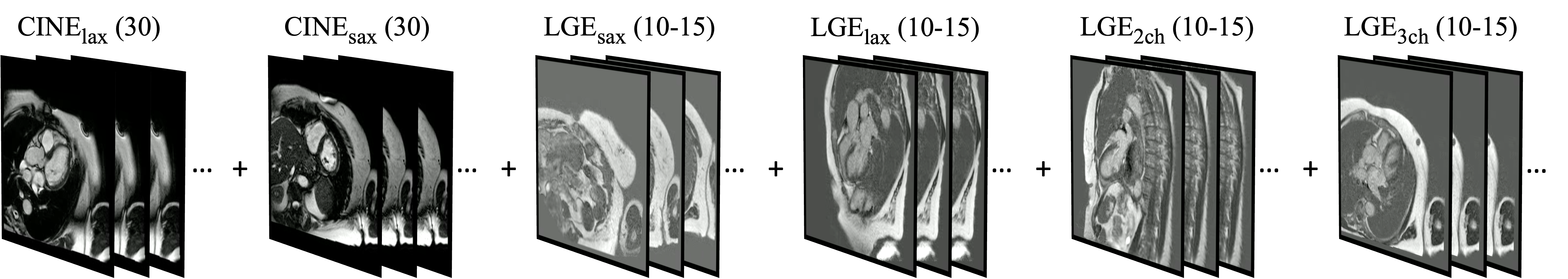}
\caption{Example of CMR image sequences constructed  by CINE\textsubscript{lax-sax} + LGE\textsubscript{lax-sax-2ch-3ch}, where ($\cdot$) represents the number of images of each type-view combination. For CINE\textsubscript{lax-sax}, each frame represents the time dimension. For LGE\textsubscript{sax}, each frame corresponds to the depth dimension, and for LGE\textsubscript{lax-2ch-3ch}, each image is duplicated to be consistent with LGE\textsubscript{sax}.
}
\label{fig:video_data}
\vspace{-10pt}
\end{figure}

\paragraph{CMR image types and views} 
In this work, we define the different settings as IMAGE-TYPE$_{view}$, where IMAGE-TYPE is the type of the image, and ${view}$ is the view of the image. Multiple views of the same image type are joined with ``-''. For example, CINE\textsubscript{lax-sax} refers to the long-axis and short-axis view of CINE. 
An example of video constructed using CINE\textsubscript{lax-sax} + LGE\textsubscript{lax-sax-2ch-3ch} is shown in Figure~\ref{fig:video_data}.

\paragraph{Tasks} We include two tasks in the experiments: (1) Retrieval, which includes text-to-video retrieval and video-to-text retrieval. Text-to-video retrieval means retrieving relevant CMR sequences based on a given textual report. Video-to-text retrieval means retrieving relevant textual reports based on a given CMR sequence. (2) Classification, which uses the embeddings from the visual encoder to carry out disease classification on the labeled datasets, including our Cardiomyopathies dataset and public ACDC dataset.

\paragraph{Evaluation metrics} For the retrieval task, we report the recall at K (R@K) metric, K = $\{5, 10, 50\}$, where RSUM is defined as the sum of recall metrics at K = $\{5, 10, 50\}$ of both video and text retrieval tasks.  For the classification task, we use the standard classification accuracy (Acc), AUC, and F1 score as evaluation metrics. 

\paragraph{Training details} For data splitting of the CMR dataset, we used the 80\%/20\% split, resulting in 11,028 studies as the training set, and  2,758 studies as the testing set. The model was pretrained on WebVid-2M dataset \citep{Bain2021FrozenIT}, which contains 2.5M video-text pairs. The detail of model parameters is shown in Table~\ref{table:hyperparam} in the Appendix~\ref{sec:Appendix-parameters}. Our model was trained on $4\times \text{NVIDIA A100}$ GPUs.

\subsection{Experimental Results on the Retrieval Task}

\begin{table}[t]
\centering
\caption{Experimental results for retrieval experiments. ($\cdot$) represents the number of input frames. Zero-Shot evaluation was done using CINE\textsubscript{lax-sax} + LGE\textsubscript{lax-sax-2ch-3ch}.}
\vspace{-10pt}
\begin{center}
\begin{adjustbox}{width=0.99\linewidth}

\begin{tabular}{l|ccc|ccc|r}
\toprule 
\multirow{2}{*}{Method} & \multicolumn{3}{c}{Text-to-Video Retrieval} & \multicolumn{3}{c}{Video-to-Text Retrieval}  \\
& R@5 & R@10 & R@50  &R@5 & R@10 & R@50  &RSUM \\
\midrule
Zero-shot (16) & 0.3 & 0.4 & 1.8 & 0.2 & 0.4 & 1.5 & 4.6\\ 
CINE\textsubscript{sax} (8) & 9.4 & 15.0 & 38.6 & 9.2 & 14.9 & 39.1 & 126.2\\
CINE\textsubscript{lax-sax} (8) & 13.9 & 21.1 & 45.2 & 13.3 & 19.9 & 44.3 & 157.7\\
LGE\textsubscript{lax-sax-2ch-3ch} (8) & 14.1 & 22.3 & 50.3 & 14.2 & 22.3 & 50.8 & 174.1 \\
CINE\textsubscript{lax-sax} + LGE\textsubscript{lax-sax} (16) & 16.4 & 23.9 & 54.0 & 15.4 & 23.9 & 54.6 & 188.1\\
CINE\textsubscript{lax-sax} + LGE\textsubscript{lax-sax-2ch-3ch} (1) & 6.3 & 9.7 & 27.0 & 6.3 & 9.6 & 27.6 & 86.7\\
CINE\textsubscript{lax-sax} + LGE\textsubscript{lax-sax-2ch-3ch} (4) & 14.5 & 21.8 & 46.7 & 14.0 & 21.8 & 45.3 & 164.0\\
CINE\textsubscript{lax-sax} + LGE\textsubscript{lax-sax-2ch-3ch} (8) & 14.8 & 23.7 & 51.0 & 14.4 & 23.4 & 51.1 & 178.5\\
CINE\textsubscript{lax-sax} + LGE\textsubscript{lax-sax-2ch-3ch} (16) & 17.9 & 25.9 & 53.1 & 17.3 & 26.0 & 54.1 & 194.3\\
CINE\textsubscript{lax-sax} + LGE\textsubscript{lax-sax-2ch-3ch} (32) & 17.7 & 26.5 & 55.3 & 17.8 & 26.1 & 56.2 & 199.8\\
CINE\textsubscript{lax-sax} + LGE\textsubscript{lax-sax-2ch-3ch} (64) & \textbf{18.5} & \textbf{28.1} & \textbf{56.3} & \textbf{18.1} & \textbf{27.5} & \textbf{56.4} & \textbf{204.8} \\
\bottomrule
\end{tabular}
\end{adjustbox}
\end{center}
\label{table:exp_results}
\vspace{-10pt}
\end{table}

\paragraph{Learned representations showed better performance than zero-shot results} 
Table~\ref{table:exp_results} shows the retrieval results. 
CMRformer demonstrates remarkable retrieval performance across a wide range of CMR data formats and all metrics, as shown in Table~\ref{table:exp_results}. Moreover, our model outperforms the zero-shot setting, which uses a model trained solely on WebVid-2M without our CMR dataset. This comparison highlights the effectiveness of our learning approach in improving the model's performance.
Overall, we found that CINE\textsubscript{lax-sax} + LGE\textsubscript{lax-sax-2ch-3ch} exhibited the best performance compared to other CMR image type/view combinations.

\paragraph{More types/views contribute better performance} 
The performance of single image types, namely CINE\textsubscript{lax-sax} and LGE\textsubscript{lax-sax-2ch-3ch}, was observed to be lower in comparison to that of multiple image types/views. CINE offers a dynamic view of the heart's motion over time, while LGE captures the distribution of fibrosis in the heart. When both image types are included in the radiology report, it enhances the semantic alignment between image sequences and text reports. The inclusion of \texttt{2ch} and \texttt{3ch} images provides more information about areas surrounding the aortic and mitral valves, respectively, in addition to the information provided by \texttt{lax} and \texttt{sax}. This inclusion of left heart and valvular views enables better differentiation of certain diseases typically diagnosed using CMR, such as hypertrophic obstructive cardiomyopathy.

\paragraph{Increasing the number of CMR images can result in better performance} 
\citet{miech_howto100m_2019} found that the number of input frames is crucial for the performance of retrieval systems. Our study's findings indicate that increasing the number of input CMR images can enhance retrieval performance. Clinically, having more frames improves the ability to capture minute adverse changes to cardiac function during different points in the cardiac cycle \citep{backhaus_defining_2021}. Our study revealed that the CINE\textsubscript{lax-sax} + LGE\textsubscript{lax-sax-2ch-3ch} (64) outperformed the [1,4,8,16,32] settings. However, larger input sizes require more computational resources and higher machine requirements, such as memory.

\begin{table}[t]
\centering
\caption{Linear probing results on the Cardiomyopathies dataset for downstream disease classification task.}
\begin{center}
\begin{adjustbox}{width=0.8\linewidth}

\begin{tabular}{l|ccc|ccc}
\toprule 
\multirow{2}{*}{Model} & \multicolumn{3}{c}{\textbf{NICM}} & \multicolumn{3}{c}{\textbf{ICM}} \\
 & Acc & AUC & F1 & Acc & AUC & F1  \\
 \midrule
Zero-shot & 0.69 & 0.69 & 0.71 & 0.77 & 0.62 & 0.41\\
SimCLR & 0.71 & 0.71 & 0.74 & 0.75 & 0.62 & 0.40\\ 
CINE\textsubscript{sax} (8) & 0.75 & 0.75 & 0.77 & 0.79 & 0.71 & 0.55 \\
CINE\textsubscript{lax-sax} (8) & 0.80 & 0.80 & 0.83 & \textbf{0.84} & 0.76 & 0.64 \\
LGE\textsubscript{lax-sax-2ch-3ch} (8) & 0.81 & 0.81 & 0.82 & \textbf{0.84} & \textbf{0.79} & \textbf{0.67}  \\
CINE\textsubscript{lax-sax} + LGE\textsubscript{lax-sax-2ch-3ch} (16) & 0.82 & 0.82 & \textbf{0.84} & 0.84 & 0.77 & 0.65 \\
CINE\textsubscript{lax-sax} + LGE\textsubscript{lax-sax-2ch-3ch} (64) & \textbf{0.84} & \textbf{0.84} & \textbf{0.85} & \textbf{0.84} & \textbf{0.79} & \textbf{0.67} \\
 \midrule
  \midrule
\multirow{2}{*}{Model} & \multicolumn{3}{c}{\textbf{AMYL}} & \multicolumn{3}{c}{\textbf{HCM}}\\
 & Acc & AUC & F1 & Acc & AUC & F1 \\
 \midrule
Zero-shot & 0.93 & 0.70 & 0.43 & 0.90 & 0.79 & 0.66 \\
SimCLR & 0.93 & 0.69 & 0.43 & 0.94 & 0.84 & 0.78\\
CINE\textsubscript{sax} (8) & 0.93 & 0.75 & 0.50 & 0.96 & 0.93 & 0.87 \\
CINE\textsubscript{lax-sax} (8)  & 0.96 & 0.81 & 0.66 & 0.98 & 0.97 & 0.94 \\
LGE\textsubscript{lax-sax-2ch-3ch} (8)  & 0.96 & 0.84 & 0.70 & 0.98 & 0.98 & 0.94 \\
CINE\textsubscript{lax-sax} + LGE\textsubscript{lax-sax-2ch-3ch} (16)  & 0.95 & 0.80 & 0.63 & \textbf{0.99} & \textbf{0.99} & \textbf{0.97} \\
CINE\textsubscript{lax-sax} + LGE\textsubscript{lax-sax-2ch-3ch} (64)  & \textbf{0.97} & \textbf{0.86} & \textbf{0.76} & \textbf{0.99} & \textbf{0.99} & \textbf{0.97} \\
\bottomrule
\end{tabular}
\end{adjustbox}
\end{center}
\label{table:lp_results}
\vspace{-10pt}
\end{table}

\subsection{Experimental Results on Image Classification Task on our Cardiomyopathies Dataset}\label{sec:classification-Cardiomyopathies}

\paragraph{Cardiomyopathies dataset} The Cardiomyopathies dataset is used for studying the clinical utility of CMR for the diagnosis and prognosis of various cardiomyopathies. There are in total of 1,939 studies included in this dataset, where 1,119 studies of ischemic cardiomyopathy (ICM), 268 cardiac amyloidosis (AMYL), 318 hypertrophic cardiomyopathy (HCM), and 1,357 studies of undifferentiated non-ischemic cardiomyopathy (NICM). 
For more demographics of the Cardiomyopathies dataset, the age range is $57.96\pm14.92$, and within the 1,939 patients, 671 are female and 1,268 are male.
The final diagnosis was identified through a chart review of all available clinical data, not just using the radiology reports. Clinical fellows were tasked with identification according to the relevant clinical guidelines. A level 3 board-certified cardiologist reviewed the results for accuracy. 

In order to assess the applicability of our trained model to downstream image classification tasks, we utilized the visual embeddings learned from the visual encoder in the CMRformer and performed linear probing. The training and testing sets for our labeled Cardiomyopathies dataset were split into 70\% and 30\%, respectively. In our Cardiomyopathies dataset, the number of positive and negative samples were 1049 and 890 for NICM, 461 and 1478 for ICM, 161 and 1778 for AMYL, and 268 and 1671 for HCM. The results of the image classification are presented in Table~\ref{table:lp_results}. It was observed that the model pretrained on WebVid-2M (zero-shot) did not generalize to CMR without fine-tuning on the CMR data. In addition, although SimCLR \citep{Chen2020ASF}, a pretrained vision model, has been proven useful in other medical imaging modalities, it did not yield appreciably better results. Our CMRformer achieved significantly better results, demonstrating that the visual embeddings learned by our model can also be useful in downstream image classification tasks.

We have observed a correlation between the linear probing performance and the retrieval performance, suggesting that the trained models have learned valuable CMR representations that can be transferred to downstream tasks. Among the four diseases of interest, linear probing achieves the highest performance on HCM and the lowest on ICM. The superior performance on HCM is expected due to the distinctive morphological differences for HCM patients, such as significantly thickened myocardium. In contrast, NICM, ICM, and AMYL are more challenging to classify. Specifically, AMYL is frequently classified as a subset of NICM, with the main differentiation being that the pathological processes and treatment for AMYL are better established compared to undifferentiated NICM. ICM can also be challenging to diagnose, as there may be a mild disease or focal disease. Furthermore, in clinical practice, the diagnosis of NICM and ICM is often unclear. Many cases of NICM may have characteristics similar to ICM, such as a focal scar in the absence of coronary obstruction, resulting in partially correct misclassification.

\subsection{Experimental Results on Image Classification Task on ACDC Dataset}
\begin{figure}[t]
\centering
\begin{minipage}[t]{.48\linewidth}
         \centering
         \includegraphics[width=0.99\textwidth]{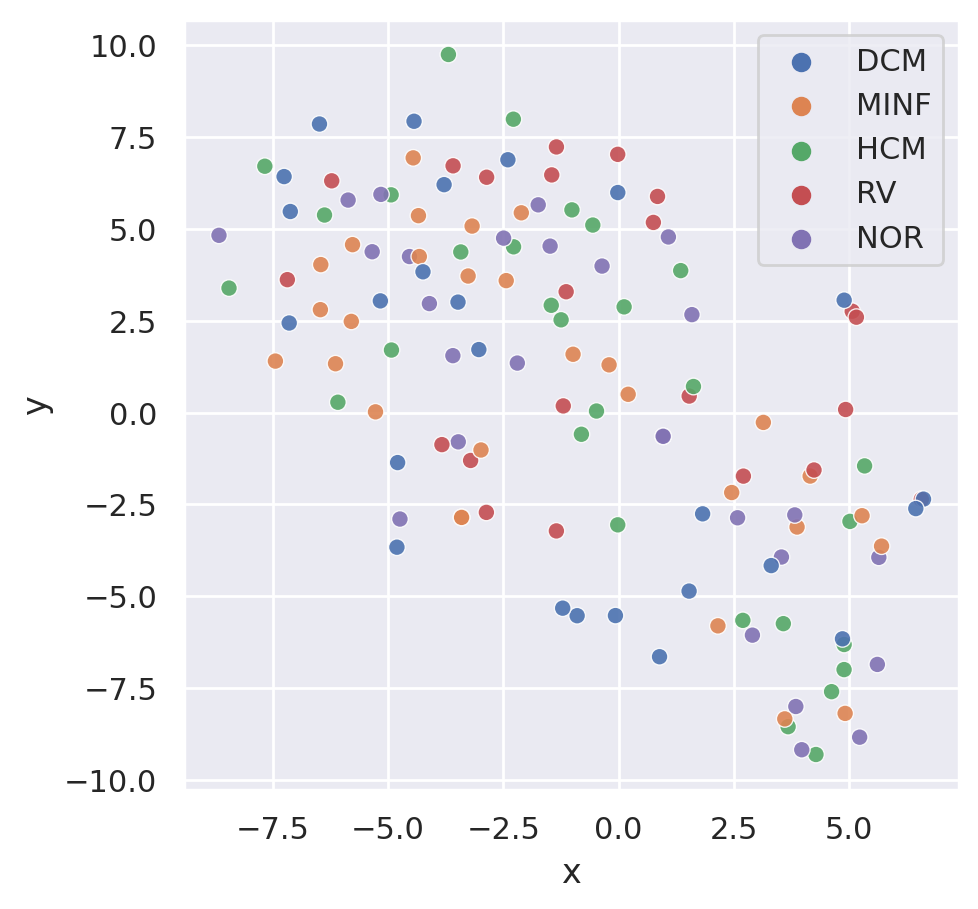}
         \caption{t-SNE visualization of zero-shot visual embeddings on the ACDC dataset.}
         \vspace{-10pt}
        \label{fig:tsne_acdc_zero_shot}
\end{minipage}
\hfill
\begin{minipage}[t]{.48\linewidth}
\centering
         \includegraphics[width=0.95\textwidth]{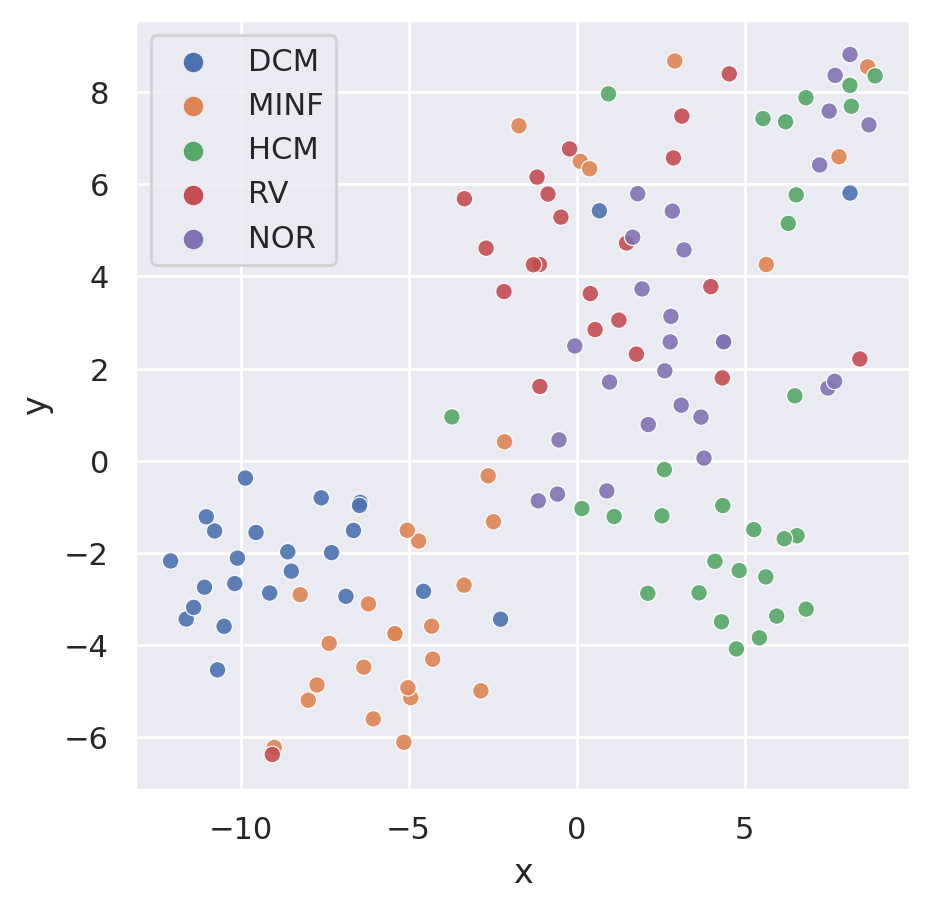}
         \caption{t-SNE visualization of learned visual embedding by CMRformer on ACDC.}
         \vspace{-10pt}
     \label{fig:tsne_acdc}
\end{minipage}
\end{figure}

To assess the generalizability of our model, we performed additional experiments on the public ACDC dataset, which only includes CINE\textsubscript{sax} data. To ensure a fair comparison, we compared our model trained on CINE\textsubscript{sax} with SimCLR \citep{Chen2020ASF} and the zero-shot setting. The multi-class classification results are presented in Table~\ref{table:lp_results_acdc}. Our model outperformed SimCLR, demonstrating its superior generalizability. Our CMRformer's learned embeddings also proved to be more beneficial in downstream image classification tasks compared to zero-shot results.
In order to present more intuitive outcomes, we have employed t-SNE \citep{van2008visualizing} to visualize the visual embedding. The findings are demonstrated in Figure~\ref{fig:tsne_acdc_zero_shot} and Figure~\ref{fig:tsne_acdc} for zero-shot setting and trained by CMRformer, respectively. Based on the figures, we observed that the visual embeddings of different classes in the zero-shot setting are intermingled. Conversely, the visual embeddings obtained from our CMRformer are more categorically separated, which elucidates why our approach delivers better image classification outcomes.

\begin{table*}[t]
\centering
\caption{Comparison of the image classification results on the ACDC dataset, where AUC is computed in a one-vs-rest manner and both F1 score and AUC are micro-averaged.}
\begin{center}
\begin{adjustbox}{width=0.45\linewidth}
\begin{tabular}{l|ccc}
\toprule 
Model & Acc & AUC & F1\\
 \midrule
Zero-shot & 0.30 & 0.59 & 0.30 \\
SimCLR & 0.42 & 0.82 & 0.42 \\
Ours (CINE\textsubscript{sax}) (8) & \textbf{0.70} & \textbf{0.95} & \textbf{0.70}  \\
\bottomrule
\end{tabular}
\end{adjustbox}
\end{center}
\label{table:lp_results_acdc}
\vspace{-10pt}
\end{table*}

\section{Discussion} 

In this work, we proposed the first multimodal vision-language contrastive learning framework, which enables the acquisition of Cardiovascular Magnetic Resonance (CMR) representations accompanied by associated cardiologist's reports. Through leveraging this framework, the acquired representations exhibit potential utility in diverse clinical contexts, ranging from the creation of robust retrieval systems to the advancement of disease classification. Our work lays the foundation for future investigations exploring the integration of multimodal learning approaches in medical imaging, which may lead to more accurate diagnoses and improved patient outcomes.

\paragraph{Different CMR image types and views} 
We conducted extensive experiments to investigate the performance of various types and views of CMR images. Our findings revealed that the performance of single image types, such as CINE\textsubscript{lax-sax} and LGE\textsubscript{lax-sax-2ch-3ch}, was lower when compared to multiple image types/views. Furthermore, the inclusion of \texttt{2ch} and \texttt{3ch} images provided additional information about the areas surrounding the aortic and mitral valves, respectively, in addition to the information already provided by \texttt{lax} and \texttt{sax}. This additional information allowed for better differentiation of diseases, such as hypertrophic obstructive cardiomyopathy.

\paragraph{Generalizability of our approach} 
Generalizability is a crucial consideration, particularly in the clinical field. To evaluate the suitability of our trained model for downstream image classification tasks, we utilized the visual embeddings acquired from the visual encoder in the CMRformer and conducted linear probing. We observed a positive correlation between the linear probing performance and the retrieval performance, indicating that our trained models have learned valuable CMR representations that can be transferred to downstream tasks. Additionally, we performed further experiments on the public ACDC dataset and demonstrated that our approach's generalizability is significantly better than the baseline methods.

\paragraph{Advancement of video-text setting} 
Previous studies \citep{zhang_contrastive_2022, huang_gloria_2021} have focused on static 2D images, but the true value of clinical imaging lies in 3D, 4D, and sometimes even 5D data. While various data fusion strategies \citep{suk_hierarchical_2014} exist to combine data from multiple independent frames, they are not easily adaptable to multi-modal pretraining. Our work proposes a method to learn from complex structured clinical image sequences and associated reports without requiring significant preprocessing. Instead of training individual encoders for each view and image type, the entire study can be processed simultaneously. Furthermore, this approach provides a straightforward way to incorporate other image types into the training process.

\paragraph{Difficulties in learning CMR} 
There are three key factors that make interpreting CMR challenging. First, it requires synthesizing information from both a single frame, such as identifying focal scar on LGE, and motion from a series of frames, such as identifying contractile dysfunction on CINE. Second, CMR patients often have multiple co-morbidities, which contribute to difficulties in identifying the clear cause for clinical symptoms. This ambiguity can lead to variability in interpretation, further misaligning image-text pairs. Finally, there are practical barriers reducing access to CMR, leading to greater than 10-fold less volume compared to other popular clinical modalities. All these issues combined make CMR one of the most difficult settings for the application of machine learning methodologies.

\paragraph{Limitations} 
Despite considerable efforts to organize the data, the data volume pales in comparison to data from other domains. The complexity of the CMR data, consisting of hundreds of images from diverse angles and types, and collected using different scanners, may render the current model inadequate or insufficiently effective in capturing and processing all the valuable information contained in the CMR image sequences. Furthermore, the limited access to public data makes it challenging to evaluate the model's generalizability comprehensively. Therefore, further experimentation using multi-center and multi-disease datasets is preferred if more public data becomes available.

\paragraph{Acknowledgements} 
This work was supported in part by the Charles and Loraine Moore Endowed Chair in Cardiovascular Imaging.
We also like to thank the Software/Data Science team in the Cleveland Clinic Imaging Informatics group for helping us acquire the images.
This project was also partially supported by funding from the Defense Advanced Research Projects Agency ADAPTER program. 

\bibliography{reference} 

\begin{thebibliography}{75}
\providecommand{\natexlab}[1]{#1}
\providecommand{\url}[1]{\texttt{#1}}
\expandafter\ifx\csname urlstyle\endcsname\relax
  \providecommand{\doi}[1]{doi: #1}\else
  \providecommand{\doi}{doi: \begingroup \urlstyle{rm}\Url}\fi

\bibitem[Alabdaljabar et~al.(2023)Alabdaljabar, Hasan, Noseworthy, Maalouf,
  Ammash, and Hashmi]{Alabdaljabar2023MachineLI}
Mohamad~S. Alabdaljabar, Babar Hasan, Peter~A. Noseworthy, Joseph Maalouf,
  Naser~M. Ammash, and Shahrukh~K Hashmi.
\newblock Machine learning in cardiology: A potential real-world solution in
  low- and middle-income countries.
\newblock \emph{Journal of Multidisciplinary Healthcare}, 16:\penalty0 285 --
  295, 2023.

\bibitem[Alayrac et~al.(2020)Alayrac, Recasens, Schneider, Arandjelovi{\'c},
  Ramapuram, De~Fauw, Smaira, Dieleman, and Zisserman]{alayrac2020self}
Jean-Baptiste Alayrac, Adri{\`a} Recasens, Rosalia Schneider, Relja
  Arandjelovi{\'c}, Jason Ramapuram, Jeffrey De~Fauw, Lucas Smaira, Sander
  Dieleman, and Andrew Zisserman.
\newblock Self-supervised multimodal versatile networks.
\newblock In \emph{NeurIPS}, 2020.

\bibitem[Azizi et~al.(2021)Azizi, Mustafa, Ryan, Beaver, Freyberg, Deaton, Loh,
  Karthikesalingam, Kornblith, and Chen]{azizi_big_2021}
Shekoofeh Azizi, Basil Mustafa, Fiona Ryan, Zachary Beaver, Jan Freyberg,
  Jonathan Deaton, Aaron Loh, Alan Karthikesalingam, Simon Kornblith, and Ting
  Chen.
\newblock Big self-supervised models advance medical image classification.
\newblock In \emph{Proceedings of the {IEEE}/{CVF} {International} {Conference}
  on {Computer} {Vision}}, pages 3478--3488, 2021.

\bibitem[Backhaus et~al.(2021)Backhaus, Metschies, Billing, Schmidt-Rimpler,
  Kowallick, Gertz, Lapinskas, Pieske-Kraigher, Pieske, and
  Lotz]{backhaus_defining_2021}
Sören~J. Backhaus, Georg Metschies, Marcus Billing, Jonas Schmidt-Rimpler,
  Johannes~T. Kowallick, Roman~J. Gertz, Tomas Lapinskas, Elisabeth
  Pieske-Kraigher, Burkert Pieske, and Joachim Lotz.
\newblock Defining the optimal temporal and spatial resolution for
  cardiovascular magnetic resonance imaging feature tracking.
\newblock \emph{Journal of Cardiovascular Magnetic Resonance}, 23\penalty0
  (1):\penalty0 1--12, 2021.
\newblock ISBN: 1532-429X Publisher: BioMed Central.

\bibitem[Bai et~al.(2019)Bai, Chen, Tarroni, Duan, Guitton, Petersen, Guo,
  Matthews, and Rueckert]{bai_self-supervised_2019}
Wenjia Bai, Chen Chen, Giacomo Tarroni, Jinming Duan, Florian Guitton,
  Steffen~E. Petersen, Yike Guo, Paul~M. Matthews, and Daniel Rueckert.
\newblock Self-{Supervised} {Learning} for {Cardiac} {MR} {Image}
  {Segmentation} by {Anatomical} {Position} {Prediction}.
\newblock pages 541--549. Springer International Publishing, 2019.
\newblock ISBN 978-3-030-32245-8.

\bibitem[Bain et~al.(2021{\natexlab{a}})Bain, Nagrani, Varol, and
  Zisserman]{Bain2021FrozenIT}
Max Bain, Arsha Nagrani, G{\"u}l Varol, and Andrew Zisserman.
\newblock Frozen in time: A joint video and image encoder for end-to-end
  retrieval.
\newblock \emph{2021 IEEE/CVF International Conference on Computer Vision
  (ICCV)}, pages 1708--1718, 2021{\natexlab{a}}.

\bibitem[Bain et~al.(2021{\natexlab{b}})Bain, Nagrani, Varol, and
  Zisserman]{bain_frozen_2021}
Max Bain, Arsha Nagrani, Gül Varol, and Andrew Zisserman.
\newblock Frozen in time: {A} joint video and image encoder for end-to-end
  retrieval.
\newblock In \emph{Proceedings of the {IEEE}/{CVF} {International} {Conference}
  on {Computer} {Vision}}, pages 1728--1738, 2021{\natexlab{b}}.

\bibitem[Bao et~al.(2019)Bao, Qiu, Tang, Zheng, and Lu]{Bao2019InvestigatingSD}
Lan-Qing Bao, Jielin Qiu, Hao Tang, Wei-Long Zheng, and Bao-Liang Lu.
\newblock Investigating sex differences in classification of five emotions from
  eeg and eye movement signals.
\newblock \emph{2019 41st Annual International Conference of the IEEE
  Engineering in Medicine and Biology Society (EMBC)}, pages 6746--6749, 2019.

\bibitem[Bernard et~al.(2018)Bernard, Lalande, Zotti, Cervenansky, Yang, Heng,
  Cetin, Lekadir, Camara, Ballester, et~al.]{bernard2018deep}
Olivier Bernard, Alain Lalande, Clement Zotti, Frederick Cervenansky, Xin Yang,
  Pheng-Ann Heng, Irem Cetin, Karim Lekadir, Oscar Camara, Miguel
  Angel~Gonzalez Ballester, et~al.
\newblock Deep learning techniques for automatic mri cardiac multi-structures
  segmentation and diagnosis: is the problem solved?
\newblock \emph{IEEE transactions on medical imaging}, 37\penalty0
  (11):\penalty0 2514--2525, 2018.

\bibitem[Bertasius et~al.(2021)Bertasius, Wang, and
  Torresani]{bertasius2021spacetime}
Gedas Bertasius, Heng Wang, and Lorenzo Torresani.
\newblock Is space-time attention all you need for video understanding?
\newblock \emph{arXiv:2102.05095}, 2021.

\bibitem[Bodenreider(2004)]{bodenreider_unified_2004}
Olivier Bodenreider.
\newblock The unified medical language system ({UMLS}): integrating biomedical
  terminology.
\newblock \emph{Nucleic acids research}, 32\penalty0 (suppl\_1):\penalty0
  D267--D270, 2004.
\newblock ISBN: 0305-1048 Publisher: Oxford University Press.

\bibitem[Buch et~al.(2022)Buch, Eyzaguirre, Gaidon, Wu, Fei-Fei, and
  Niebles]{buch_revisiting_2022}
Shyamal Buch, Cristóbal Eyzaguirre, Adrien Gaidon, Jiajun Wu, Li~Fei-Fei, and
  Juan~Carlos Niebles.
\newblock Revisiting the" video" in video-language understanding.
\newblock In \emph{Proceedings of the {IEEE}/{CVF} {Conference} on {Computer}
  {Vision} and {Pattern} {Recognition}}, pages 2917--2927, 2022.

\bibitem[Cai et~al.(2019)Cai, Wang, Li, and Liu]{Cai2019ASO}
Qiong Cai, Hao Wang, Zhenmin Li, and Xinyu Liu.
\newblock A survey on multimodal data-driven smart healthcare systems:
  Approaches and applications.
\newblock \emph{IEEE Access}, 7:\penalty0 133583--133599, 2019.

\bibitem[Chen et~al.(2022)Chen, Bhopalwala, Dewaswala, Arunachalam, Enayati,
  Farahani, Pasupathy, Lokineni, Bos, Noseworthy, Arsanjani, Erickson, Geske,
  Ackerman, Araoz, and Arruda-Olson]{Chen2022DeepNN}
David Chen, Huzefa~M. Bhopalwala, Nakeya Dewaswala, Shivaram~Poigai
  Arunachalam, Moein Enayati, Nasibeh~Zanjirani Farahani, Kalyan~Sunder
  Pasupathy, Sravani Lokineni, J.~Martijn Bos, Peter~A. Noseworthy, Reza
  Arsanjani, Bradley~James Erickson, Jeffrey~B. Geske, Michael~J. Ackerman,
  Philip~A. Araoz, and Adelaide~M. Arruda-Olson.
\newblock Deep neural network for cardiac magnetic resonance image
  segmentation.
\newblock \emph{Journal of Imaging}, 8, 2022.

\bibitem[Chen et~al.(2019)Chen, Bentley, Mori, Misawa, Fujiwara, and
  Rueckert]{chen_self-supervised_2019}
Liang Chen, Paul Bentley, Kensaku Mori, Kazunari Misawa, Michitaka Fujiwara,
  and Daniel Rueckert.
\newblock Self-supervised learning for medical image analysis using image
  context restoration.
\newblock \emph{Medical Image Analysis}, 58:\penalty0 101539, 2019.
\newblock ISSN 1361-8415.
\newblock \doi{https://doi.org/10.1016/j.media.2019.101539}.

\bibitem[Chen et~al.(2020{\natexlab{a}})Chen, Kornblith, Norouzi, and
  Hinton]{Chen2020ASF}
Ting Chen, Simon Kornblith, Mohammad Norouzi, and Geoffrey~E. Hinton.
\newblock A simple framework for contrastive learning of visual
  representations.
\newblock \emph{ArXiv}, abs/2002.05709, 2020{\natexlab{a}}.

\bibitem[Chen et~al.(2020{\natexlab{b}})Chen, Li, Yu, Kholy, Ahmed, Gan, Cheng,
  and Liu]{Chen2020UNITERUI}
Yen-Chun Chen, Linjie Li, Licheng Yu, Ahmed~El Kholy, Faisal Ahmed, Zhe Gan,
  Yu~Cheng, and Jingjing Liu.
\newblock Uniter: Universal image-text representation learning.
\newblock In \emph{ECCV}, 2020{\natexlab{b}}.

\bibitem[Dash et~al.(2019)Dash, Shakyawar, Sharma, and Kaushik]{dash_big_2019}
Sabyasachi Dash, Sushil~Kumar Shakyawar, Mohit Sharma, and Sandeep Kaushik.
\newblock Big data in healthcare: management, analysis and future prospects.
\newblock \emph{Journal of Big Data}, 6\penalty0 (1):\penalty0 54, June 2019.
\newblock ISSN 2196-1115.
\newblock \doi{10.1186/s40537-019-0217-0}.

\bibitem[Devlin et~al.(2019)Devlin, Chang, Lee, and Toutanova]{devlin2019bert}
J.~Devlin, Ming-Wei Chang, Kenton Lee, and Kristina Toutanova.
\newblock {BERT}: Pre-training of deep bidirectional transformers for language
  understanding.
\newblock In \emph{NAACL-HLT}, 2019.

\bibitem[Dewaswala et~al.(2022)Dewaswala, Chen, Bhopalwala, Kaggal, Murphy,
  Bos, Geske, Gersh, Ommen, Araoz, Ackerman, and
  Arruda-Olson]{Dewaswala2022NaturalLP}
Nakeya Dewaswala, David~C. Chen, Huzefa~M. Bhopalwala, Vinod~C Kaggal, Sean~P.
  Murphy, J.~Martijn Bos, Jeffrey~B. Geske, Bernard~J. Gersh, Steve~R Ommen,
  Philip~A. Araoz, Michael~J. Ackerman, and Adelaide~M. Arruda-Olson.
\newblock Natural language processing for identification of hypertrophic
  cardiomyopathy patients from cardiac magnetic resonance reports.
\newblock \emph{BMC Medical Informatics and Decision Making}, 22, 2022.

\bibitem[Dou et~al.(2021)Dou, Xu, Gan, Wang, Wang, Wang, Zhu, Peng, Liu, and
  Zeng]{Dou2021AnES}
Zi-Yi Dou, Yichong Xu, Zhe Gan, Jianfeng Wang, Shuohang Wang, Lijuan Wang,
  Chenguang Zhu, Nanyun Peng, Zicheng Liu, and Michael Zeng.
\newblock An empirical study of training end-to-end vision-and-language
  transformers.
\newblock \emph{ArXiv}, abs/2111.02387, 2021.

\bibitem[Fonseca et~al.(2011)Fonseca, Backhaus, Bluemke, Britten, Chung, Cowan,
  Dinov, Finn, Hunter, Kadish, et~al.]{fonseca2011cardiac}
Carissa~G Fonseca, Michael Backhaus, David~A Bluemke, Randall~D Britten, Jae~Do
  Chung, Brett~R Cowan, Ivo~D Dinov, J~Paul Finn, Peter~J Hunter, Alan~H
  Kadish, et~al.
\newblock The cardiac atlas project—an imaging database for computational
  modeling and statistical atlases of the heart.
\newblock \emph{Bioinformatics}, 27\penalty0 (16):\penalty0 2288--2295, 2011.

\bibitem[Gabeur et~al.(2020)Gabeur, Sun, Alahari, and Schmid]{gabeur2020multi}
Valentin Gabeur, Chen Sun, Karteek Alahari, and Cordelia Schmid.
\newblock Multi-modal transformer for video retrieval.
\newblock In \emph{ECCV}, 2020.

\bibitem[Gan et~al.(2020)Gan, Chen, Li, Zhu, Cheng, and
  Liu]{Gan2020LargeScaleAT}
Zhe Gan, Yen-Chun Chen, Linjie Li, Chen Zhu, Yu~Cheng, and Jingjing Liu.
\newblock Large-scale adversarial training for vision-and-language
  representation learning.
\newblock \emph{ArXiv}, abs/2006.06195, 2020.

\bibitem[Good and Su(2013)]{good_crowdsourcing_2013}
Benjamin~M. Good and Andrew~I. Su.
\newblock Crowdsourcing for bioinformatics.
\newblock \emph{Bioinformatics}, 29\penalty0 (16):\penalty0 1925--1933, 2013.
\newblock ISBN: 1460-2059 Publisher: Oxford University Press.

\bibitem[Hamilton(2015)]{kaggle_dsbcc}
Booz~Allen Hamilton.
\newblock Kaggle 2nd annual data science bowl cardiac challenge.
\newblock
  \url{https://www.kaggle.com/competitions/second-annual-data-science-bowl},
  2015.

\bibitem[Hayat et~al.(2022)Hayat, Geras, and Shamout]{Hayat2022MedFuseMF}
Nasir Hayat, Krzysztof~J. Geras, and Farah~E. Shamout.
\newblock Medfuse: Multi-modal fusion with clinical time-series data and chest
  x-ray images.
\newblock \emph{ArXiv}, abs/2207.07027, 2022.

\bibitem[He et~al.(2022)He, Chen, Xie, Li, Dollár, and
  Girshick]{he_masked_2022}
Kaiming He, Xinlei Chen, Saining Xie, Yanghao Li, Piotr Dollár, and Ross
  Girshick.
\newblock Masked autoencoders are scalable vision learners.
\newblock In \emph{Proceedings of the {IEEE}/{CVF} {Conference} on {Computer}
  {Vision} and {Pattern} {Recognition}}, pages 16000--16009, 2022.

\bibitem[Hosny et~al.(2018)Hosny, Parmar, Quackenbush, Schwartz, and
  Aerts]{hosny_artificial_2018}
Ahmed Hosny, Chintan Parmar, John Quackenbush, Lawrence~H. Schwartz, and
  Hugo~JWL Aerts.
\newblock Artificial intelligence in radiology.
\newblock \emph{Nature Reviews Cancer}, 18\penalty0 (8):\penalty0 500--510,
  2018.
\newblock ISBN: 1474-175X Publisher: Nature Publishing Group UK London.

\bibitem[Huang et~al.(2021)Huang, Shen, Lungren, and Yeung]{huang_gloria_2021}
Shih-Cheng Huang, Liyue Shen, Matthew~P. Lungren, and Serena Yeung.
\newblock Gloria: {A} multimodal global-local representation learning framework
  for label-efficient medical image recognition.
\newblock In \emph{International {Conference} on {Computer} {Vision}}, pages
  3942--3951, 2021.

\bibitem[Irvin et~al.(2019)Irvin, Rajpurkar, Ko, Yu, Ciurea-Ilcus, Chute,
  Marklund, Haghgoo, Ball, and Shpanskaya]{irvin_chexpert_2019}
Jeremy Irvin, Pranav Rajpurkar, Michael Ko, Yifan Yu, Silviana Ciurea-Ilcus,
  Chris Chute, Henrik Marklund, Behzad Haghgoo, Robyn Ball, and Katie
  Shpanskaya.
\newblock Chexpert: {A} large chest radiograph dataset with uncertainty labels
  and expert comparison.
\newblock In \emph{Proceedings of the {AAAI} conference on artificial
  intelligence}, volume~33, pages 590--597, 2019.
\newblock ISBN 2374-3468.
\newblock Issue: 01.

\bibitem[Johnson et~al.(2019)Johnson, Pollard, Berkowitz, Greenbaum, Lungren,
  Deng, Mark, and Horng]{johnson_mimic-cxr_2019}
Alistair~EW Johnson, Tom~J. Pollard, Seth~J. Berkowitz, Nathaniel~R. Greenbaum,
  Matthew~P. Lungren, Chih-ying Deng, Roger~G. Mark, and Steven Horng.
\newblock {MIMIC}-{CXR}, a de-identified publicly available database of chest
  radiographs with free-text reports.
\newblock \emph{Scientific data}, 6\penalty0 (1):\penalty0 317, 2019.
\newblock ISBN: 2052-4463 Publisher: Nature Publishing Group UK London.

\bibitem[Kahn~Jr et~al.(2009)Kahn~Jr, Langlotz, Burnside, Carrino, Channin,
  Hovsepian, and Rubin]{kahn2009toward}
Charles~E Kahn~Jr, Curtis~P Langlotz, Elizabeth~S Burnside, John~A Carrino,
  David~S Channin, David~M Hovsepian, and Daniel~L Rubin.
\newblock Toward best practices in radiology reporting.
\newblock \emph{Radiology}, 252\penalty0 (3):\penalty0 852--856, 2009.

\bibitem[Kim et~al.(2021)Kim, Son, and Kim]{Kim2021ViLTVT}
Wonjae Kim, Bokyung Son, and Ildoo Kim.
\newblock Vilt: Vision-and-language transformer without convolution or region
  supervision.
\newblock In \emph{ICML}, 2021.

\bibitem[Kline et~al.(2022)Kline, Wang, Li, Dennis, Hutch, Xu, Wang, Cheng, and
  Luo]{Kline2022MultimodalML}
Adrienne~S. Kline, Hanyin Wang, Yikuan Li, Saya Dennis, Meghan~R. Hutch,
  Zhenxing Xu, Fei Wang, Feixiong Cheng, and Yuan Luo.
\newblock Multimodal machine learning in precision health: A scoping review.
\newblock \emph{NPJ Digital Medicine}, 5, 2022.

\bibitem[Krittanawong et~al.(2020)Krittanawong, Virk, Bangalore, Wang, Johnson,
  Pinotti, Zhang, Kaplin, Narasimhan, Kitai, Baber, Halperin, and
  Tang]{Krittanawong2020MachineLP}
Chayakrit Krittanawong, Hafeez Ul~Hassan Virk, Sripal Bangalore, Zhen Wang,
  Kipp~W. Johnson, Rachel Pinotti, Hongju Zhang, Scott~L. Kaplin, Bharat
  Narasimhan, Takeshi Kitai, Usman Baber, Jonathan~L. Halperin, and
  W.~H.~Wilson Tang.
\newblock Machine learning prediction in cardiovascular diseases: a
  meta-analysis.
\newblock \emph{Scientific Reports}, 10, 2020.

\bibitem[Lee et~al.(2017)Lee, Luo, Ngiam, Zhang, Zheng, Chen, Ooi, and
  Yip]{lee_big_2017}
Chonho Lee, Zhaojing Luo, Kee~Yuan Ngiam, Meihui Zhang, Kaiping Zheng, Gang
  Chen, Beng~Chin Ooi, and Wei Luen~James Yip.
\newblock Big healthcare data analytics: {Challenges} and applications.
\newblock \emph{Handbook of large-scale distributed computing in smart
  healthcare}, pages 11--41, 2017.
\newblock ISBN: 3319582798 Publisher: Springer.

\bibitem[Lei et~al.(2021)Lei, Li, Zhou, Gan, Berg, Bansal, and
  Liu]{lei2021less}
Jie Lei, Linjie Li, Luowei Zhou, Zhe Gan, Tamara~L Berg, Mohit Bansal, and
  Jingjing Liu.
\newblock Less is more: Clipbert for video-and-language learning via sparse
  sampling.
\newblock \emph{arXiv preprint arXiv:2102.06183}, 2021.

\bibitem[Li et~al.(2021)Li, Selvaraju, Gotmare, Joty, Xiong, and
  Hoi]{Li2021AlignBF}
Junnan Li, Ramprasaath~R. Selvaraju, Akhilesh~Deepak Gotmare, Shafiq~R. Joty,
  Caiming Xiong, and Steven C.~H. Hoi.
\newblock Align before fuse: Vision and language representation learning with
  momentum distillation.
\newblock In \emph{NeurIPS}, 2021.

\bibitem[Li et~al.(2022{\natexlab{a}})Li, Li, Xiong, and Hoi]{Li2022BLIPBL}
Junnan Li, Dongxu Li, Caiming Xiong, and Steven C.~H. Hoi.
\newblock Blip: Bootstrapping language-image pre-training for unified
  vision-language understanding and generation.
\newblock In \emph{ICML}, 2022{\natexlab{a}}.

\bibitem[Li et~al.(2022{\natexlab{b}})Li, Gao, Niu, Xiao, Liu, Liu, Wu, and
  Wang]{li2022unimo}
Wei Li, Can Gao, Guocheng Niu, Xinyan Xiao, Hao Liu, Jiachen Liu, Hua Wu, and
  Haifeng Wang.
\newblock Unimo-2: End-to-end unified vision-language grounded learning.
\newblock \emph{arXiv preprint arXiv:2203.09067}, 2022{\natexlab{b}}.

\bibitem[Li et~al.(2020)Li, Yin, Li, Hu, Zhang, Zhang, Wang, Hu, Dong, Wei,
  Choi, and Gao]{Li2020OscarOA}
Xiujun Li, Xi~Yin, Chunyuan Li, Xiaowei Hu, Pengchuan Zhang, Lei Zhang, Lijuan
  Wang, Houdong Hu, Li~Dong, Furu Wei, Yejin Choi, and Jianfeng Gao.
\newblock Oscar: Object-semantics aligned pre-training for vision-language
  tasks.
\newblock In \emph{ECCV}, 2020.

\bibitem[Lin et~al.(2014)Lin, Maire, Belongie, Hays, Perona, Ramanan, Dollár,
  and Zitnick]{lin_microsoft_2014}
Tsung-Yi Lin, Michael Maire, Serge Belongie, James Hays, Pietro Perona, Deva
  Ramanan, Piotr Dollár, and C.~Lawrence Zitnick.
\newblock Microsoft coco: {Common} objects in context.
\newblock In \emph{Computer {Vision}–{ECCV} 2014: 13th {European}
  {Conference}, {Zurich}, {Switzerland}, {September} 6-12, 2014, {Proceedings},
  {Part} {V} 13}, pages 740--755. Springer, 2014.
\newblock ISBN 3-319-10601-5.

\bibitem[Liu et~al.(2021)Liu, Qiu, Zheng, and Lu]{Liu2021ComparingRP}
Wei Liu, Jielin Qiu, Wei-Long Zheng, and Bao-Liang Lu.
\newblock Comparing recognition performance and robustness of multimodal deep
  learning models for multimodal emotion recognition.
\newblock \emph{IEEE Transactions on Cognitive and Developmental Systems},
  14:\penalty0 715--729, 2021.

\bibitem[Luo and Yuille(2019)]{gst}
Chenxu Luo and Alan Yuille.
\newblock Grouped spatial-temporal aggretation for efficient action
  recognition.
\newblock In \emph{ICCV}, 2019.

\bibitem[Miech et~al.(2018)Miech, Laptev, and Sivic]{miech18learning}
Antoine Miech, Ivan Laptev, and Josef Sivic.
\newblock Learning a text-video embedding from incomplete and heterogeneous
  data.
\newblock \emph{arXiv}, 2018.

\bibitem[Miech et~al.(2019)Miech, Zhukov, Alayrac, Tapaswi, Laptev, and
  Sivic]{miech_howto100m_2019}
Antoine Miech, Dimitri Zhukov, Jean-Baptiste Alayrac, Makarand Tapaswi, Ivan
  Laptev, and Josef Sivic.
\newblock Howto100m: {Learning} a text-video embedding by watching hundred
  million narrated video clips.
\newblock In \emph{Proceedings of the {IEEE}/{CVF} {International} {Conference}
  on {Computer} {Vision}}, pages 2630--2640, 2019.

\bibitem[Miech et~al.(2020)Miech, Alayrac, Smaira, Laptev, Sivic, and
  Zisserman]{miech20endtoend}
Antoine Miech, Jean-Baptiste Alayrac, Lucas Smaira, Ivan Laptev, Josef Sivic,
  and Andrew Zisserman.
\newblock End-to-end learning of visual representations from uncurated
  instructional videos.
\newblock In \emph{CVPR}, 2020.

\bibitem[Moon et~al.(2019)Moon, Sagheb, Liu, Chen, Bos, Geske, Noseworthy,
  Ackerman, Shellum, Chaudhry, Ommen, Araoz, Nishimura, Liu, and
  Arruda-Olson]{Moon2019Abstract1A}
Sungrim Moon, Elham Sagheb, Sijia Liu, David~C. Chen, Martijn Bos, Jeffrey~B.
  Geske, Peter~A. Noseworthy, Michael~J. Ackerman, Jane~L. Shellum, Rajeev
  Chaudhry, Steve~R Ommen, Philip~A. Araoz, Rick~A. Nishimura, Hongfang Liu,
  and Adelaide~M. Arruda-Olson.
\newblock Abstract 13811: An automated natural language processing algorithm to
  classify magnetic resonance imaging reports containing positive diagnoses of
  hypertrophic cardiomyopathy.
\newblock \emph{Circulation}, 2019.

\bibitem[M{\"u}ller et~al.(2022)M{\"u}ller, Kaissis, Zou, and
  Rueckert]{muller2022radiological}
Philip M{\"u}ller, Georgios Kaissis, Congyu Zou, and Daniel Rueckert.
\newblock Radiological reports improve pre-training for localized imaging tasks
  on chest x-rays.
\newblock In \emph{Medical Image Computing and Computer Assisted
  Intervention--MICCAI 2022: 25th International Conference, Singapore,
  September 18--22, 2022, Proceedings, Part V}, pages 647--657. Springer, 2022.

\bibitem[Nagendran et~al.(2020)]{nagendran_artificial_2020}
Myura Nagendran et~al.
\newblock Artificial intelligence versus clinicians: systematic review of
  design, reporting standards, and claims of deep learning studies.
\newblock \emph{bmj}, 368, 2020.
\newblock ISBN: 1756-1833 Publisher: British Medical Journal Publishing Group.

\bibitem[Nakashima et~al.(2022)Nakashima, Jang, Basnet, Benovoy, Tang, Nguyen,
  Kwon, Hwang, and Chen]{Nakashima2022InteractionOA}
Makiya Nakashima, Inyeop Jang, Ramesh Basnet, Mitchel Benovoy, Wen Tang,
  Christopher~T. Nguyen, Deborah~H Kwon, Tae~Hyun Hwang, and David Chen.
\newblock Interaction of a priori anatomic knowledge with self-supervised
  contrastive learning in cardiac magnetic resonance imaging.
\newblock \emph{ArXiv}, abs/2205.12429, 2022.

\bibitem[Petersen et~al.(2013)Petersen, Matthews, Bamberg, Bluemke, Francis,
  Friedrich, Leeson, Nagel, Plein, and Rademakers]{petersen_imaging_2013}
Steffen~E. Petersen, Paul~M. Matthews, Fabian Bamberg, David~A. Bluemke,
  Jane~M. Francis, Matthias~G. Friedrich, Paul Leeson, Eike Nagel, Sven Plein,
  and Frank~E. Rademakers.
\newblock Imaging in population science: cardiovascular magnetic resonance in
  100,000 participants of {UK} {Biobank}-rationale, challenges and approaches.
\newblock \emph{Journal of Cardiovascular Magnetic Resonance}, 15\penalty0
  (1):\penalty0 1--10, 2013.
\newblock ISBN: 1532-429X Publisher: BioMed Central.

\bibitem[Petersen et~al.(2015)Petersen, Matthews, Francis, Robson, Zemrak,
  Boubertakh, Young, Hudson, Weale, and Garratt]{petersen_uk_2015}
Steffen~E. Petersen, Paul~M. Matthews, Jane~M. Francis, Matthew~D. Robson,
  Filip Zemrak, Redha Boubertakh, Alistair~A. Young, Sarah Hudson, Peter Weale,
  and Steve Garratt.
\newblock {UK} {Biobank}’s cardiovascular magnetic resonance protocol.
\newblock \emph{Journal of cardiovascular magnetic resonance}, 18\penalty0
  (1):\penalty0 1--7, 2015.
\newblock ISBN: 1532-429X Publisher: BioMed Central.

\bibitem[Qiu and Zhao(2018)]{Qiu2018DataEV}
Jielin Qiu and Wei-Ye Zhao.
\newblock Data encoding visualization based cognitive emotion recognition with
  ac-gan applied for denoising.
\newblock \emph{2018 IEEE 17th International Conference on Cognitive
  Informatics \& Cognitive Computing (ICCI*CC)}, pages 222--227, 2018.

\bibitem[Qiu et~al.(2022)Qiu, Zhu, Xu, Huang, Rosenberg, Weber, Liu, and
  Zhao]{Qiu2022CardiacDD}
Jielin Qiu, Jiacheng Zhu, Mengdi Xu, Peide Huang, Michael Rosenberg, Douglas
  Weber, Emerson Liu, and Ding Zhao.
\newblock Cardiac disease diagnosis on imbalanced electrocardiography data
  through optimal transport augmentation.
\newblock 2022.

\bibitem[Qiu et~al.(2023)Qiu, Han, Zhu, Xu, Rosenberg, Liu, Weber, and
  Zhao]{Qiu2023TransferKF}
Jielin Qiu, William Han, Jiacheng Zhu, Mengdi Xu, Michael Rosenberg, Emerson
  Liu, Douglas Weber, and Ding Zhao.
\newblock Transfer knowledge from natural language to electrocardiography: Can
  we detect cardiovascular disease through language models?
\newblock \emph{ArXiv}, abs/2301.09017, 2023.

\bibitem[Quer et~al.(2021)Quer, Arnaout, Henne, and Arnaout]{Quer2021MachineLA}
Giorgio Quer, Ramy~A. Arnaout, Michael Henne, and Rima Arnaout.
\newblock Machine learning and the future of cardiovascular care: Jacc
  state-of-the-art review.
\newblock \emph{Journal of the American College of Cardiology}, 77 3:\penalty0
  300--313, 2021.

\bibitem[Radford and Narasimhan(2018)]{Radford2018ImprovingLU}
Alec Radford and Karthik Narasimhan.
\newblock Improving language understanding by generative pre-training.
\newblock 2018.

\bibitem[Radford et~al.(2021)Radford, Kim, Hallacy, Ramesh, Goh, Agarwal,
  Sastry, Askell, Mishkin, and Clark]{radford_learning_2021}
Alec Radford, Jong~Wook Kim, Chris Hallacy, Aditya Ramesh, Gabriel Goh,
  Sandhini Agarwal, Girish Sastry, Amanda Askell, Pamela Mishkin, and Jack
  Clark.
\newblock Learning transferable visual models from natural language
  supervision.
\newblock In \emph{International conference on machine learning}, pages
  8748--8763. PMLR, 2021.
\newblock ISBN 2640-3498.

\bibitem[Sanh et~al.(2019)Sanh, Debut, Chaumond, and Wolf]{distilbert}
Victor Sanh, Lysandre Debut, Julien Chaumond, and Thomas Wolf.
\newblock Distilbert, a distilled version of {BERT:} smaller, faster, cheaper
  and lighter.
\newblock \emph{CoRR}, abs/1910.01108, 2019.

\bibitem[Schulz-Menger et~al.(2020)Schulz-Menger, Bluemke, Bremerich, Flamm,
  Fogel, Friedrich, Kim, von Knobelsdorff-Brenkenhoff, Kramer, Pennell, Plein,
  and Nagel]{schulz-menger_standardized_2020}
Jeanette Schulz-Menger, David~A. Bluemke, Jens Bremerich, Scott~D. Flamm,
  Mark~A. Fogel, Matthias~G. Friedrich, Raymond~J. Kim, Florian von
  Knobelsdorff-Brenkenhoff, Christopher~M. Kramer, Dudley~J. Pennell, Sven
  Plein, and Eike Nagel.
\newblock Standardized image interpretation and post-processing in
  cardiovascular magnetic resonance - 2020 update.
\newblock \emph{Journal of Cardiovascular Magnetic Resonance}, 22\penalty0
  (1):\penalty0 19, March 2020.
\newblock ISSN 1532-429X.
\newblock \doi{10.1186/s12968-020-00610-6}.

\bibitem[Suk et~al.(2014)Suk, Lee, and Shen]{suk_hierarchical_2014}
Heung-Il Suk, Seong-Whan Lee, and Dinggang Shen.
\newblock Hierarchical feature representation and multimodal fusion with deep
  learning for {AD}/{MCI} diagnosis.
\newblock \emph{NeuroImage}, 101:\penalty0 569--582, 2014.
\newblock ISSN 1053-8119.
\newblock \doi{10.1016/j.neuroimage.2014.06.077}.

\bibitem[Suo et~al.(2019)Suo, Zhong, Ma, Yuan, Gao, and Zhang]{Suo2019MetricLO}
Qiuling Suo, Weida Zhong, Fenglong Ma, Ye~Yuan, Jing Gao, and Aidong Zhang.
\newblock Metric learning on healthcare data with incomplete modalities.
\newblock In \emph{International Joint Conference on Artificial Intelligence},
  2019.

\bibitem[Van Der~Maaten(2008)]{van2008visualizing}
Laurens~JP Van Der~Maaten.
\newblock Visualizing high-dimensional data using t-sne.
\newblock \emph{J Mach Learn Res}, 9:\penalty0 2579, 2008.

\bibitem[{Wang} et~al.(2019){Wang}, {Xiong}, {Wang}, {Qiao}, {Lin}, {Tang}, and
  {Van Gool}]{wang_tsn}
L.~{Wang}, Y.~{Xiong}, Z.~{Wang}, Y.~{Qiao}, D.~{Lin}, X.~{Tang}, and L.~{Van
  Gool}.
\newblock Temporal segment networks for action recognition in videos.
\newblock \emph{IEEE Transactions on Pattern Analysis and Machine
  Intelligence}, 41\penalty0 (11):\penalty0 2740--2755, 2019.
\newblock \doi{10.1109/TPAMI.2018.2868668}.

\bibitem[Wang et~al.(2017)Wang, Peng, Lu, Lu, Bagheri, and
  Summers]{wang_chestx-ray8_2017}
Xiaosong Wang, Yifan Peng, Le~Lu, Zhiyong Lu, Mohammadhadi Bagheri, and
  Ronald~M. Summers.
\newblock Chestx-ray8: {Hospital}-scale chest x-ray database and benchmarks on
  weakly-supervised classification and localization of common thorax diseases.
\newblock pages 2097--2106, 2017.

\bibitem[Wang et~al.(2022)Wang, Wu, Agarwal, and Sun]{wang_medclip_2022}
Zifeng Wang, Zhenbang Wu, Dinesh Agarwal, and Jimeng Sun.
\newblock Medclip: {Contrastive} learning from unpaired medical images and
  text.
\newblock \emph{arXiv preprint arXiv:2210.10163}, 2022.

\bibitem[Windsor et~al.(2023)Windsor, Jamaludin, Kadir, and
  Zisserman]{windsor2023vision}
Rhydian Windsor, Amir Jamaludin, Timor Kadir, and Andrew Zisserman.
\newblock Vision-language modelling for radiological imaging and reports in the
  low data regime.
\newblock \emph{arXiv preprint arXiv:2303.17644}, 2023.

\bibitem[Xu et~al.(2021)Xu, Ghosh, Huang, Okhonko, Aghajanyan, Metze,
  Zettlemoyer, and Feichtenhofer]{xu_videoclip_2021}
Hu~Xu, Gargi Ghosh, Po-Yao Huang, Dmytro Okhonko, Armen Aghajanyan, Florian
  Metze, Luke Zettlemoyer, and Christoph Feichtenhofer.
\newblock Videoclip: {Contrastive} pre-training for zero-shot video-text
  understanding.
\newblock \emph{arXiv preprint arXiv:2109.14084}, 2021.

\bibitem[Yan and Pei(2022)]{yan2022clinical}
Bin Yan and Mingtao Pei.
\newblock Clinical-bert: Vision-language pre-training for radiograph diagnosis
  and reports generation.
\newblock In \emph{Proceedings of the AAAI Conference on Artificial
  Intelligence}, volume~36, pages 2982--2990, 2022.

\bibitem[Yan et~al.(2023)Yan, Li, Marques, Gao, and Fong]{Yan2023ARO}
Keyue Yan, Tengyue Li, Jo{\~a}o Alexandre~L{\^o}bo Marques, Juntao Gao, and
  Simon~James Fong.
\newblock A review on multimodal machine learning in medical diagnostics.
\newblock \emph{Mathematical Biosciences and Engineering}, 2023.

\bibitem[Zhai and Wu(2019)]{Zhai2019ClassificationIA}
Andrew Zhai and Hao-Yu Wu.
\newblock Classification is a strong baseline for deep metric learning.
\newblock In \emph{BMVC}, 2019.

\bibitem[Zhang et~al.(2022)Zhang, Jiang, Miura, Manning, and
  Langlotz]{zhang_contrastive_2022}
Yuhao Zhang, Hang Jiang, Yasuhide Miura, Christopher~D. Manning, and Curtis~P.
  Langlotz.
\newblock Contrastive learning of medical visual representations from paired
  images and text.
\newblock In \emph{Machine {Learning} for {Healthcare} {Conference}}, pages
  2--25. PMLR, 2022.
\newblock ISBN 2640-3498.

\bibitem[Zhu et~al.(2022)Zhu, Qiu, Yang, Weber, Rosenberg, Liu, Li, and
  Zhao]{Zhu2022GeoECGDA}
Jiacheng Zhu, Jielin Qiu, Zhuolin Yang, Douglas Weber, Michael~A. Rosenberg,
  Emerson Liu, Bo~Li, and Ding Zhao.
\newblock Geoecg: Data augmentation via wasserstein geodesic perturbation for
  robust electrocardiogram prediction.
\newblock In \emph{Machine Learning in Health Care}, 2022.

\end{thebibliography}

\newpage
\appendix
\section{Experimental Parameters}\label{sec:Appendix-parameters}

The parameters are shown in Table~\ref{table:hyperparam}. 

\begin{table*}[htp]
\centering
\caption{Model parameters in the experiments.}
\vspace{5pt}
\begin{center}
\begin{adjustbox}{width=0.99\linewidth}
\begin{tabular}{c|c}
\toprule 
Hyperparameter & Value \\
\midrule
n\_gpu & 4 \\
optimizer & Adam \\
lr & 3e-5 \\
loss & NormSoftmaxLoss \\
epoch & 100 \\
batch\_size & 16 \\
extraction\_res & 256 \\
input\_res & 224 \\ 
stride & 1 \\
\bottomrule
\end{tabular}
\hspace{0.5cm}
\begin{tabular}{c|c}
\toprule 
video params & Value \\
\midrule
model & SpaceTimeTransformer \\
arch\_config & base\_patch16\_224 \\
num\_frames & \{1, 4, 8, 16, 32, 64\} \\
pretrained & true \\
time\_init & zeros \\
\midrule
text params & Value \\
\midrule
model & distilbert-base-uncased \\
pretrained & true \\
\bottomrule
\end{tabular}

\end{adjustbox}
\end{center}
\label{table:hyperparam}
\end{table*}

\section{More Introduction about Existing Datasets}\label{sec:Appendix-datasets}

\paragraph{ACDC dataset} The Automated Cardiac Diagnosis Challenge (ACDC) dataset \citep{bernard2018deep} is a public dataset comprising 150 clinical CMRs acquired at the University Hospital of Dijon, France on either a 1.5T Siemens Area and 3.0T Siemens Trio scanner. The dataset includes only CINE\textsubscript{sax} images. We used the pathology labels included within the dataset, which include myocardial infarction with systolic heart failure, dilated cardiomyopathy, HCM, abnormal right ventricle, and normal. Classes are evenly distributed (30 in each class), and the dataset has pre-determined training (100) and test (50) sets. 

\paragraph{DSB-CC dataset} The 2nd Annual Data Bowl by Kaggle and Booz Allen Hamilton \citep{kaggle_dsbcc} included 1,140 CMR studies. The dataset was limited to only CINE\textsubscript{sax} images and did not include any disease labels or segmentation labels. Instead, the dataset provided only numeric biomarker measurements, which typically require physician segmentation to acquire. This limits the inherent value of this dataset for pretraining purposes. 

\paragraph{Comparison with UK Biobank} The UK Biobank contains a large, semi-public CMR dataset representing UK citizens being prospectively followed \citep{petersen_imaging_2013,petersen_uk_2015}. Although significantly larger than our data, the UK Biobank is not a good representation of clinical imaging taken within the standard of care. Rather, it is a prospective registry of UK citizens with no specific disease focus and, therefore, is biased heavily toward healthy individuals. The large subset (5000) of the study has been heavily analyzed for imaging biomarkers, and the images can be connected to a multitude of other follow-up data. However, the studies are not interpreted by a radiologist. Furthermore, the CMR studies are collected at four sites with a purposely designed protocol and single-model MR machine making its generalizability to general practice questionable.

\section{More Related Work}

\paragraph{Machine learning in cardiology}
With the development of machine learning (ML), ML approaches have been widely explored in many applications, including cardiology. 
\citet{Moon2019Abstract1A} proposed to autonomously classify MRI narrative reports containing positive diagnoses of HCM using natural language processing (NLP) techniques.
\citet{Krittanawong2020MachineLP} proposed a composite of the predictive ability of ML algorithms of coronary artery disease, heart failure, stroke, and cardiac arrhythmias. 
\citet{Quer2021MachineLA} reviewed the ML methods that have been applied in cardiology to help clinicians to understand. 
\citet{Qiu2022CardiacDD} used a Multi-Feature
Transformer model to classify ECG signals for disease detection.  
\citet{Chen2022DeepNN} proposed a multi-image type UNet for cardiac magnetic resonance (CMR) segmentation. 
\citet{Zhu2022GeoECGDA} proposed a data augmentation method via Wasserstein geodesic perturbation for robust ECG prediction. 
\citet{Nakashima2022InteractionOA} used a segmentation network to explicitly local the heart in CMR images, followed by self-supervised contrastive learning in multiple diagnostic tasks.
\citet{Qiu2023TransferKF} explored that the learned knowledge by Large Language Models (LLMs) can be transferred to clinical cardiology EEG, for disease section and diagnosis report generation.
\citet{Alabdaljabar2023MachineLI} reviewed what was known about AI/ML in cardiovascular medicine, and discussed how it could benefit low- and middle-income countries (LMICs).

\paragraph{Multimodal learning in healthcare applications}

Multimodal learning techniques have also been proven to be useful in many healthcare applications. 
\citet{Suo2019MetricLO} proposed a metric learning framework to perform missing modality completion and multi-modal metric learning simultaneously for healthcare data. 
\citet{Cai2019ASO} provided a comprehensive survey of existing techniques on multimodal data-driven smart healthcare systems. 
\citet{Qiu2018DataEV} visualized EEG signals into images for affective computing.
\citet{Bao2019InvestigatingSD} investigated the sex difference in classifying  emotions from EEG and eye movement signals.
\citet{Liu2021ComparingRP} compared the recognition performance and robustness of two multimodal models on the emotion recognition problem. 
\citet{Dewaswala2022NaturalLP} used the NLP method to accurately extract the diagnosis of HCM from CMR reports. 
\citet{Hayat2022MedFuseMF} proposed a multi-modal fusion method for clinical time-series data and chest X-ray images.
\citet{Kline2022MultimodalML} summarized the current studies in multimodal learning in precision health.
\citet{Yan2023ARO}  reviewed and summarized several recent papers that deal with multimodal machine learning in disease detection, and identified some topics for future research directions.

\end{document}